\documentclass[journal]{IEEEtai}

\usepackage[colorlinks,urlcolor=blue,linkcolor=blue,citecolor=blue]{hyperref}

\usepackage{color,array}
\usepackage{bbm}
\usepackage{arcs}
\usepackage{graphicx}
\usepackage{algorithm}
\usepackage{algorithmic}
\usepackage{color}
\usepackage{amsmath,amssymb}
\usepackage{url}
\usepackage{comment}
\usepackage{subfloat}
\usepackage{subfig} 
\usepackage{float} 
\usepackage{chngpage}

\newcommand{\Saracomm}{\textcolor[rgb]{0,0,0}}
\newcommand{\giac}{\textcolor[rgb]{0,0,0}}
\newcommand{\Saracommdue}{\textcolor[rgb]{0,0,0}}
\newcommand{\Sararev}{\textcolor[rgb]{0,0,0}}

\usepackage{xspace}
\usepackage{soul}
\usepackage{color}
\usepackage[usenames,dvipsnames]{xcolor}
\newcommand{\hlc}[2][yellow]{ {\sethlcolor{#1} \hl{#2}} }

\begin{document}

\title{Rule-based out-of-distribution detection} 

\author{Giacomo De Bernardi, Sara Narteni, Enrico Cambiaso and Maurizio Mongelli \IEEEmembership{Member, IEEE}
\thanks{The authors are with the National Research Council of Italy (CNR) - Institute of Electronics and Information and Telecommunications Engineering (IEIIT), Corso F. M. Perrone 24, 16152, Genoa, Italy (e-mail: name.surname@ieiit.cnr.it)}
\thanks{G.De Bernardi is also with Università degli studi di Genova, The Electrical, Electronics and Telecommunication Engineering and Naval Architecture Department (DITEN), Genova, Italy}
\thanks{S.Narteni is also with Politecnico di Torino, Department of Control and Computer Engineering (DAUIN), 10129, Turin, Italy}
}

\markboth{Journal of IEEE Transactions on Artificial Intelligence, Vol. 00, No. 0, Month 2020}
{De Bernardi G. \MakeLowercase{\textit{et al.}}: Rule-based Out-Of-Distribution Detection}

\maketitle

\begin{abstract}
Out-of-distribution detection is one of the most critical issue in the deployment of machine learning. The data analyst must assure that data in operation should be compliant with the training phase as well as understand if the environment has changed in a way that autonomous decisions would not be safe anymore.
The method of the paper is based on eXplainable Artificial Intelligence (XAI); it takes into account different metrics to identify any resemblance between in-distribution and out of, as seen by the XAI model. The approach is non-parametric and distributional assumption free. The validation over complex scenarios (predictive maintenance, vehicle platooning, covert channels in cybersecurity) corroborates both precision in detection and evaluation of training-operation conditions proximity. 
\end{abstract}

\begin{IEEEImpStatement}
Many sectors these days address safe AI: automotive (SOTIF), avionics (SAE
G-34/EUROCAE WG-114), ISO/IEC (JTC 1/SC 42) and healthcare. Safe AI means understanding under which conditions autonomous actuations may lead to hazards. The impact of research here is to make AI aware of this, thus understanding under which conditions it may operate without detrimental effect to the human or the environment. Examples may involve the prevention of: dangerous manoeuvres by autonomous cars, inaccurate clinical diagnosis by artificial doctors, wrong decision making in cyberwarfare and in many other sectors (energy, finance). The theoretical analysis here is empowered by computational and incremental groupwise analysis in order to increase the readiness level of the proposed approach. 
\end{IEEEImpStatement}

\begin{IEEEkeywords}
Out-of-distribution detection, eXplainable AI, mutual information, open data.
\end{IEEEkeywords}

\section{Notation and List of Acronyms}
\begin{center}
\footnotesize
\begin{tabular}{|c|c| }\hline
 OoD &  Out of distribution\\ \hline
 ODD &  OoD detection\\\hline
 ML &  Machine learning\\ \hline
XAI & eXplainable Artificial Intelligence\\ \hline
 $TR$ & Training set\\ \hline
 $OP$ & Operational set\\ \hline
 $tr_i$& $i$-th training subset\\ \hline
 $op_i$ & $i$-th operational subset\\ \hline
 $n_s$ & number of data samples in a split\\ \hline
 $N_{r}$ & Number of rules \\ \hline
 $N_{tr}$ & Number of training splits\\ \hline
 $N_{op}$ & Number of operational splits\\ \hline
 $\mathcal{R}_{tr}$ & Training reference ruleset\\\hline
 $r_i$ &$i$-th rule\\ \hline
 $h_{i}^j$ & $j$-th hit for the $i$-th rule\\ \hline
 ${l_p}$ & ${l_p}$ norm\\ \hline
 $\mu I$& Mutual information\\ \hline
 $W\mu I$& Weighted mutual information\\ \hline
 $RB I$& Rule based information\\ \hline
 $H$ & Entropy\\ \hline
\end{tabular}
\end{center}
\newpage
\section{Introduction}\label{sec:intro}

\IEEEPARstart{T}{he} problem of out-of-distribution (OoD) detection (ODD) deals with comparing the working conditions of a machine learning model with those considered during the training process. The comparison is performed at the operational level to understand if the new data belong to a probability distribution different from that driving the data collection of the training phase. In case of divergence between training and operation, the system must generate an alarm because the performance of the model may no longer conform to what was measured at the training stage (even in case of successfully passed generalization tests\footnote{Generalization bounds, see, e.g., \cite{Mammarella}, concern the gap that exists between the empirical risk, calculated on the data actually available (on which the model is trained) and the theoretical risk, calculated on the distribution of probability that represents the data; this probability distribution is unknown in closed-form and, in the ODD context, represents the "in-distribution".}).
The problem represents a very important challenge for the secure application of machine learning\Sararev{, and is fundamental in the context of trustworthy AI \cite{nascita2023improving,dondio2011trust}}. The recent standards in avionics \cite{EASAfirstdraft,EASA}, automotive \cite{SOTIF, AutomotiveCornerCases} and ISO/IEC, as well as other regulatory initiatives in medical informatics \cite{assessmentAIbio}, pose the problem of identifying all those operating conditions that can have an impact on safety. 

\begin{figure}
    \centering
    \includegraphics[width=0.5\textwidth]{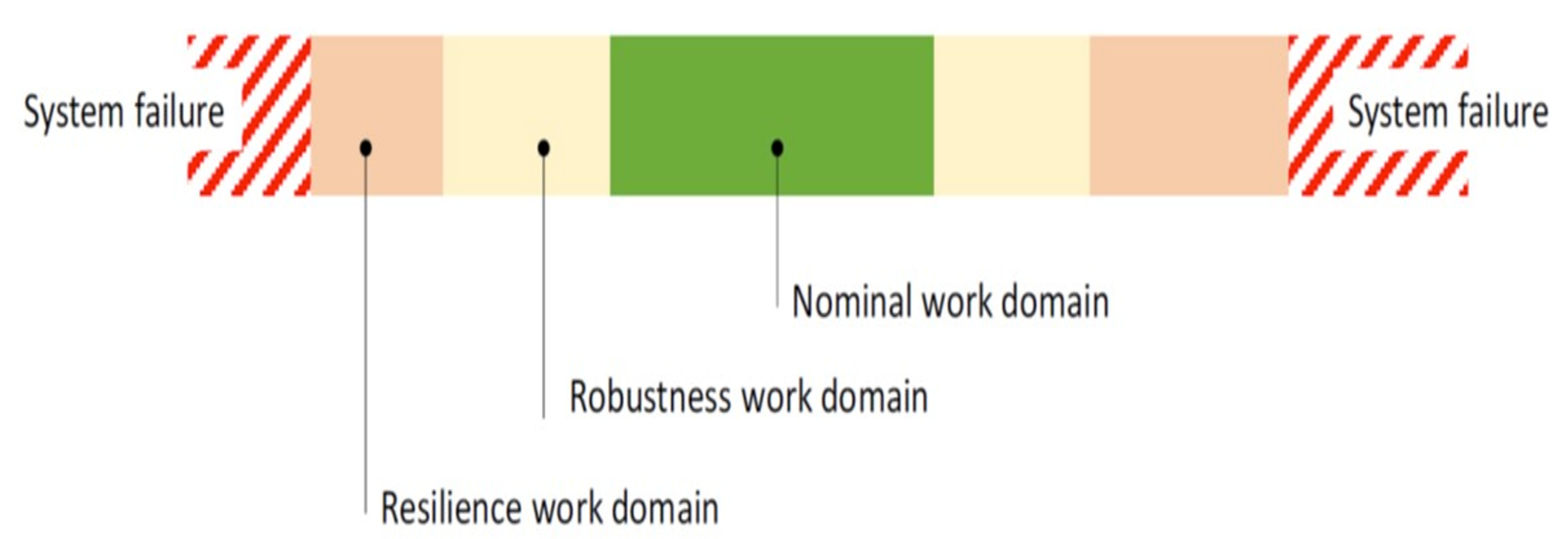}
    \caption{Illustration of the work domains as reported in \cite{EASAfirstdraft}. From central \textcolor[rgb]{0,0.7,0}{ green} bar to side \textcolor[rgb]{1,0.8,0}{yellow}/\textcolor[rgb]{1,0.7,0.4}{ orange}/{\color{red} red} bars, the nominal domain shifts and the severity of the OoD increases in parallel.}
    \label{fig:easa}
\end{figure}

The EASA Fig. \ref{fig:easa} shows the different levels of severity of the OoD on operational data. The \textcolor[rgb]{0,0.7,0}{ green} bar denotes compliance with training data (in-distribution), the \textcolor[rgb]{1,0.8,0}{yellow} color reflects an OoD zone where the autonomous function still produces accurate indications; \textcolor[rgb]{1,0.7,0.4}{ orange} color indicates an OoD area where the autonomous function is fallacious, but the system does not degenerate into dangerous conditions (the surrounding conditions of the environment are still compatible with safe actuations), while \textcolor[rgb]{1,0.0,0.0} {red} signals that the system may fall into dangerous conditions (if driven by the autonomous function). The tests of autonomous safety-critical actuation should include all the conditions in the mentioned color gradations, at least by simulation analysis. 
Although the literature in the field of OoD already poses solutions based on labelled data or through anomaly detection, as evidenced by \cite{SSD, KNN}, the OoD according to distributional assumption-free and OoD-agnostic criteria is still an open problem\footnote{Distributional assumption-free means no closed-form expressions of in- and out- probability distributions are considered. OoD-agnostic means no information about ODD conditions is considered. Another important issue, which is related with the assumption of probabilistic Gaussian or mixed-Gaussian functions, is to avoid calculating the covariance matrix from data, which can be numerically unstable.}.  

\subsection{Contribution}
The proposed method is designed under these criteria with the added advantage of avoiding any parameter tuning. It is based on the evaluation of the histogram generated by the frequency of validation of a rule-based model by the data themselves. The histogram generated during the training phase represents a fingerprint to be verified at runtime. If the data at runtime generates a histogram ``significantly different'' from the training one, it means that the data are OoD. Unlike K-NN \cite{KNN} and Neural Networks distance \cite{NNdistance}, where a single distance criterion is defined, the similarity measure can be derived through multiple metrics. This offers support to the tests mentioned by EASA, since the proposed method measures incremental cases of departure from in-distribution. \giac{In addition this is a groupwise method which makes the ODD more robust as outlined in Sec \ref{sec:groupwise}.}

\section{Related work}
ODD has become an important theme in ML field, since the recognition of unseen data either ``similar'' or not (in- or out) to the ones the ML system has been trained on may lead to potentially fatal consequences; indeed, a system should not only correctly classify what is known, yet also and most importantly should recognise what is not known for action to be taken.

Most of the solutions proposed to address the problem of the OoD make strong distributional assumptions of the feature space \cite{Lee} or suppose they are given a training in and out probability density function (pdf) \cite{OOD}, but this not always holds in practice. What is more, lots of statistic tests fail to estimate the real distribution of training data (data are not enough and the pdf are too coarse) \cite{KNN}. \giac{There are lots of widespread supervised methods used across OoD detection: model-based methods such as the ODIN \cite{ODIN}, distance-based methods like OODL\cite{abdelzad}, density-based methods as the energy-based OoD Detection method \cite{Energy} and some threshold-based methods, including Maximum Softmax Probability \cite{Softmax} or Autoencoder \cite{autoencodernn}. Other approaches use outlier detection methods as the Isolation Forest \cite{Isolation_Forest} and label shift in deep learning is also considered in \cite{Vince}. 
 Under distributional assumptions free hypothesis, unsupervised OoD detection methods need the right tuning of some parameters as \cite{KNN} and \cite{Autoencoder}. Starting from eXplainability, our solution still maintains the former and does not rely on any critical parameter setting.}

\section{Logic Learning Machine}\label{sec:llm}
\Sararev{The rule-based model adopted in our work is called Logic Learning Machine (LLM), an efficient implementation of Switching Neural Networks \cite{SNN}, developed and available in Rulex software platform \footnote{\url{https://www.rulex.ai}}.
\noindent However, we remark that the methods for OoD detection presented in the paper can be easily extended to any other kind of rule-based model, such as decision tree or tree ensembles like  random forests or Skope-Rules \footnote{\url{https://github.com/scikit-learn-contrib/skope-rules}}.} \Sararev{\noindent Given some input data, the LLM provides a classification model represented by ruleset $\mathcal{R} = \{r_k\}_{k=1,\ldots,N_r}$, with each rule $r_k$ expressed through the following structure: \textcolor{blue}{\textbf{if}} $<$\textit{premise}$>$ \textcolor{blue}{\textbf{then}} $<$\textit{consequence}$>$.
The $<$\textit{premise}$>$ is made up of the logical conjunction (AND) of conditions on the input features and the $<$\textit{consequence}$>$ constitutes the output of the classification rule.} \Sararev{\noindent Rule generation process occurs in three steps. First, a discretization and binarization of the feature space is performed by using the inverse-only-one coding. The resulting binary strings are then concatenated into a single large string representing the considered samples. Shadow clustering is then used to build logical structures, called implicants, which are finally transformed into simple conditions and combined into a collection of intelligible rules \cite{llmappl1,Sara}.}

\section{Rule-based ODD}\label{sec:method}

\subsection{Rule hits histograms}\label{subsec:isto}

\Saracomm{Let us denote with $\mathcal{R}_{tr}$ a set of rules generated from a training set and let $N_r$ be the number of rules composing it.}
Let $N_{tr}$ and $N_{op}$ be the numbers of splits of the training domain and the operational one, respectively, and let $N_h = N_{tr} + N_{op}$ be the total number of splits. \Saracomm{Let $n_s$ be the number of data samples present in a split.}
For each split, samples \footnote{Samples can satisfy multiple rules and there may be operational samples satisfying none of the rules.} may (or not) satisfy each rule a certain number of times. We refer to this number as the \textit{number of hits} for that rule.
Therefore we define $N_h$ vectors, \Saracomm{considering this number scaled by the split size $n_s$}:
\begin{equation}
\footnotesize
    \mathbf{h}^j = \big\{h_{i}^j\big\}, \;\;h_{i}^j\in[0,1] ,\;i = 1,\ldots,N_r,\; j =1,\ldots,N_h\; 
\label{eq:histograms}
\end{equation}

Each vector $\mathbf{h}^j$ can be thought as a histogram.

\Sararev{Rule hits are the key starting point of the proposed methods. Therefore, particular care should be posed on the quality of the training reference ruleset generating them. For example, feature reduction methods on the input data may provide a simplified ruleset, thus more interpretable. Nevertheless, it would result in less informative rule hits. That is, the less variables we include in the model, the more general would be the support of the resulting rules, which would be more likely to be frequently verified by the data samples. As a result, histograms shape would be flattened and the methods for OoD detection would be less performing.}

\subsection{Data splits}
At training stage, we exploit $N_{tr}$ splits of the dataset $TR = \{tr_1, \ldots, tr_{N_{tr}}\}$. \Saracomm{These splits become the baseline for building the in-distribution histograms, as per Eq. \ref{eq:histograms}, representing the numbers of hits obtained by testing the rules in $\mathcal{R}_{tr}$ on each considered split.} 
Two different algorithms are then studied, based on the data organization in operation: $N_{op}=1$ when only one split is available and $N_{op}>1 $ with more than one split. The first case is suitable for data scarcity in operation and simplifies the calculations.

\subsection{Adopted Metrics}\label{subsec:metrics}
The metrics driving ODD are as follows:
\begin{itemize}
\setlength\itemsep{-1em}
    \item Weighted mutual information $W\mu I$, used when only one operational split is available ($N_{op}=1$), as described in Sec. \ref{subsec:firstscenario} \\
    \item Rule-based information $RB I$, used when the operational data are sufficient to perform multiple splits ($N_{op}>1$), as per Sec. \ref{subsec:secondscenario} \\
    \item $l_p$ norm, with $p=1,2$; their computation is performed in the same way for both scenarios ($N_{op} \geq 1$)\\
\end{itemize}
The order of the hits with respect to the rules drives modification to canonical mutual information, $W\mu I$ and $RB I$, as explained in Appendix \ref{subsec:RULE}.
For all metrics, the idea is to compare values computed in operation with the ranges achieved in training (\textit{baseline}). 
\Sararev{Specifically, we expect that histograms generated at training or operational stages have different shapes, providing an indication of OoD. Thus, through the Algorithms presented in the following Sections, we try to quantify such a behavior. Algorithms  \ref{algo:weight_train}-\ref{algo:weighted_operational} (for $N_{op} = 1$) and \ref{algo:RBI_training}-\ref{algo:RBI_operational} (for $N_{op} > 1$) both share the same methodological approach. As per the training part, the reference ruleset $\mathcal{R}_{tr}$ is applied to the training data splits to retrieve the rule hits histograms; then the metrics of interest are computed over couples of training histograms and a \emph{baseline} is defined. At operational, rule hits histograms are derived for the couples formed by training and operational data split(s) and the values of the metrics are computed as well. 
Eventually, an ODD is acknowledged whenever the largest portion of operational values falls outside the training baseline for at least one of the metrics (i.e., through  minority voting of the metrics).
We finally remark that such methodologies are designed for working at runtime; therefore, adding any pre-processing module, e.g. any instance selection, would make the whole process slower, which would not be always acceptable when dealing with safety-critical scenarios.}

\subsection{First scenario: $N_{op}=1$} \label{subsec:firstscenario}
\subsubsection{Training setting}
The first scenario deals with a single operational split $op_1$. We first present the procedure for the training domain and then for the operational one. 
As to Eq. \ref{eq:histograms}, the training matrix-like structure shown in Table \ref{tab:hits_matrix_alg_1_train} consequently arises. 

\begin{table}[h]
    \centering
    \begin{tabular}{|c|c|c|c|} \hline
         &$tr_1$ &....&$tr_{N_{tr}}$\\ \hline
        $r_1$ & $h_{1}^{tr_1}$ &... &  $h_{1}^{tr_{N_{tr}}}$ \\ \hline
        $r_2$   & $h_{2}^{tr_1}$ &... &  $h_{2}^{tr_{N_{tr}}}$\\ \hline
        $.$   &.&.&.\\ \hline
        $.$    &.&.&.\\ \hline
        $r_{N_r}$ & $h_{N_r}^{tr_1}$ &... &  $h_{N_r}^{tr_{N_{tr}}}$\\ \hline
    \end{tabular}
    \caption{Training numbers of hits table. Each column refers to a training split $tr_i \in TR$ and each row to a rule $r_i \in \mathcal{R}_{tr}$.}
    \label{tab:hits_matrix_alg_1_train}
\end{table}

Based on that table, weighted mutual information and norms are computed as described in Algorithm \ref{algo:weight_train}. 

\begin{algorithm}[h]
\caption{Weighted Mutual Information and Norms at Training Stage} \label{algo:weight_train}
\footnotesize
$i,j$=1,$\ldots$,$N_{tr}$, $i\neq j $\\
Input: Table \ref{tab:hits_matrix_alg_1_train}\\
Output: Training baselines $W\mu I_{base}$ and ${l_p}^{base}$\\
$\rule{8.5cm}{0.02cm}$\\
1a. Define the weight associated with $tr_i$ and $tr_j$, $ \alpha_{i,j}\in (0,1)$, $\forall i$, $\forall j$:
\begin{equation*}
    \alpha_{i,j}={1\over{N_r}}\sum_{r=1}^{N_r}{(|h_{r}^{tr_i}-h_{r}^{tr_j}|)}
\end{equation*}
1b. \hbox{Compute the weighted entropies $H(tr_i),\; H(tr_j),\;H(tr_i,tr_j)\;$}, $\forall i$, $\forall j$ 
\begin{equation*}
    H(tr_i)= - \sum_{r=1}^{N_r}[\alpha_{i,j} P({h_{r}^{tr_i}})\cdot log(\alpha_{i,j} P({h_{r}^{tr_i}}))]
\end{equation*}
\begin{equation*}
    H(tr_j)=- \sum_{r=1}^{N_r}[\alpha_{i,j} P({h_{r}^{tr_j}})\cdot log(\alpha_{i,j} P({h_{r}^{tr_j}}))]
\end{equation*}
\begin{equation*}
    H(tr_i,tr_j)= -\sum_{r=1}^{N_r}[\alpha_{i,j} P({h_{r}^{tr_i}},{h_{r}^{tr_j}})\cdot log(\alpha_{i,j} P({h_{r}^{tr_i}},{h_{r}^{tr_j}}))]
\end{equation*}
2. Compute the weighted mutual information ($W\mu I$): 
\begin{equation*}
W\mu I(tr_i,tr_j)= [H(tr_i)+H(tr_j)-H(tr_i,tr_j)], \forall i, \forall j
\end{equation*}
3. Compute the baseline $W\mu I_{base}$:
\begin{equation*}
\begin{split}
   W\mu I_{base}\doteq[\min_{i,j}(W\mu I(tr_i,tr_j)), \max_{i,j}(W\mu I(tr_i,tr_j))]
   \end{split}
\end{equation*}\label{eq:weightedMIrange}
4. Compute $l_p$ ($p$=1,2) norms:
\begin{equation*}
   l_p(tr_i,tr_j)=\Bigl[\sum_{r=1}^{N_r}(|h_{r}^{tr_i}-h_{r}^{tr_j}|)^p\Bigr]^{1\over p}, \forall i, \forall j 
\end{equation*}
5. Compute the baseline ${l_p}^{base}$ ($p$=1,2):
\begin{equation*}
\begin{split}
  {l_p}^{base} \doteq [\min_{i,j}(l_p(tr_i,tr_j), \max_{i,j}(l_p(tr_i,tr_j))]
\end{split}
\end{equation*}\label{eq:L1normbaseline}
\end{algorithm}

\subsubsection{Operational setting}
We now present the procedure when an operational set is considered. As to Eq. \ref{eq:histograms}, we can build the training-operational matrix as in Table \ref{tab:hits_matrix_alg_1_op}.

\begin{table}[h]
\centering
\begin{tabular}{|c|c|c|c|c|} \hline
     &$tr_1$ &....&$tr_{N_{tr}}$&$op_1$\\ \hline
    $r_1$ & $h_{1}^{tr_1}$ &... &  $h_{1}^{tr_{N_{tr}}}$&  $h_{1}^{op_1}$  \\ \hline
    $r_2$   & $h_{2}^{tr_1}$ &... &  $h_{2}^{tr_{N_{tr}}}$&  $h_{2}^{op_1}$\\ \hline
    $.$   &.&.&.&\\ \hline
    $.$    &.&.&.& \\ \hline
    $r_{N_r}$ & $h_{N_r}^{tr_1}$ &... &  $h_{N_r}^{tr_{N_{tr}}}$&  $h_{N_r}^{op_1}$ \\ \hline
\end{tabular}
\caption{Training-operational number of hits table with $N_{op}=1$. Columns refers to the training splits $tr_i \in TR$ and the operational split $op_1$, each row to a rule $r_i \in \mathcal{R}_{tr}$.}
\label{tab:hits_matrix_alg_1_op}
\end{table}

Weighted mutual information and norms are then computed as described in Algorithm \ref{algo:weighted_operational}.

\begin{algorithm}[h]
\caption{Weighted Mutual Information and Norms at Operational Stage} \label{algo:weighted_operational}
\footnotesize
$i = 1, \ldots, N_{tr}$, $p=1,2$ \\
Input: Table \ref{tab:hits_matrix_alg_1_op}; baseline ranges $W\mu I_{base}$ and ${l_p}^{base}\;$ \\
Output: ODD through  $W\mu I$ and ${l_p}\;$\\
$\rule{8cm}{0.02cm}$\\
1. Compute the weighted entropies $H(tr_i)$, $H(op_1)$ and $H(tr_i,op_1)$ as done in the Algorithm \ref{algo:weight_train} (steps 1a-1b) $\forall i$\\
2. Compute the weighted mutual information ($W\mu I$): 
\begin{equation*}
\begin{split}
    W\mu I(tr_i,op_1)=[H(tr_i)+H(op_1)-H(tr_i,op_1)],
    \forall i
\end{split}
\end{equation*}
3.Compute $l_p$ norms:
\begin{equation*}
   l_p(tr_i,op_1)=\Bigl[\sum_{r=1}^{N_r}(|h_{r}^{tr_i}-h_{r}^{op_1}|)^p\Bigr]^{1\over p}, \forall i 
\end{equation*}
4.OoD detection:
\begin{center}
\textcolor{blue}{IF}  $W\mu I(tr_i,op_1) \notin W\mu I_{base} $  \textit{ for the majority of $i$} \textcolor{blue}{THEN} flag is \textcolor{red}{$on$}\\

\textcolor{blue}{IF}  $l_p(tr_i,op_1) \notin {l_p}^{base} $  \textit{ for the majority of $i$} \textcolor{blue}{THEN}  flag is \textcolor{red}{$on$},\\
\textcolor{blue}{IF} \textit{\{at least one flag is \textcolor{red}{on}\}} \textcolor{blue}{THEN} $op_1$ is \textcolor{red}{OoD}
\end{center}
\end{algorithm}

\subsection{Second scenario: $N_{op}>1$}\label{subsec:secondscenario}

Again by following the notation of section \ref{subsec:isto}, several splits of the training domain $TR$ are defined $TR = \{tr_1, \ldots, tr_{N_{tr}}\}$, together with an analogous set for the operational domain, $OP = \{op_1, \ldots, op_{N_{op}}\}$. 
The training splits are organized in two subsets: $TR1= \{tr_1, \ldots, tr_k\}$, $TR2 = \{tr_{k+1},\ldots, tr_{N_{tr}}\}$ \Saracommdue{with $k=N_{tr}-N_{op}-1$} and, based on a leave-one-out cross-validation \cite{bishop2006pattern}, we consider $TR2_m=TR2 \setminus \{tr_m\},\; m = k+1,\ldots,N_{tr}$. These sets drive the computation of the baseline according to the \Saracomm{rule-based} information ($RB I$) (algorithm \ref{algo:RBI_training}, table \ref{tab:hits_matrix_alg_1_train} as a reference). Algorithm \ref{algo:RBI_operational} defines the inherent ODD by taking table \ref{tab:hits_matrix_operational} as a reference. \\
The estimation of the Gaussian distributions follows the maximum likelihood principle (see 2.5.1 of \cite{LibrolezioneML}). Building $N_r$ separate Gaussian distributions (one for each row of the table) allows to tackle with the curse of dimensionality problem in parameters estimation, in place of building a single, multi-dimensional Gaussian distribution for the entire table (see, e.g., 2.5.7 of \cite{LibrolezioneML}). We remark that this methodology still complies with the distributional assumption-free (as stated in Sec. \ref{sec:intro}) because the Gaussian distribution estimation concerns the rule hits and not the data and \giac{furthermore thanks to the metodology followed to construct each histogram we can take advantage of the central limit theorem and the law of large numbers.}

\begin{table}[H]
    \centering
    \begin{tabular}{|c|c|c|c|c|c|c|} \hline
         &$tr_1$ &....&$tr_{N_{tr}}$&$op_1$&...& $op_{N_{op}}$  \\ \hline
        $r_1$ & $h_{1}^{tr_1}$ &... &  $h_{1}^{tr_{N_{tr}}}$&  $h_{1}^{op_1}$ &...&  $h_{1}^{op_{N_{op}}}$ \\ \hline
        $r_2$   & $h_{2}^{tr_1}$ &... &  $h_{2}^{tr_{N_{tr}}}$&  $h_{2}^{op_1}$ &...&  $h_{2}^{op_{N_{op}}}$ \\ \hline
        $.$   &.&.&.&.&.&. \\ \hline
        $.$    &.&.&.&.&.&. \\ \hline
        $r_{N_r}$ & $h_{N_r}^{tr_1}$ &... &  $h_{N_r}^{tr_{N_{tr}}}$&  $h_{N_r}^{op_1}$ &...&  $h_{N_r}^{op_{N_{op}}}$\\ \hline
    \end{tabular}
    \caption{Training-operational number of hits table with $N_{op} > 1$. Columns $tr_1,\ldots,tr_{N_{tr}}$ refer to training splits (further organized in $TR1$ and $TR2_m$ sets, see Sec. \ref{subsec:secondscenario}), columns $op_1,\ldots,op_{N_{op}}$ are the splits in operation. Each row refers to a rule $r_i \in \mathcal{R}_{tr}$.}
    \label{tab:hits_matrix_operational}
\end{table}

\begin{algorithm}[h]
\caption{Rule-based Information at Training Stage} \label{algo:RBI_training}
\footnotesize
$i = 1, \ldots, N_{tr}, p=1,2, j = 1,\dots, N_r, tr_i \in TR2_m$;\\
Inputs: \hbox{ $TR1$ and $TR2_m,\;\; m = k+1,\ldots,N_{tr}$, \Saracommdue{$k=N_{tr}-N_{op}-1$}}\\
Output: Training baselines $RBI_{base}$ and ${l_p}^{base}$\\
$\rule{8.5cm}{0.02cm}$\\

1. \hbox{Compute $\mu^{TR1}_j = \frac{1}{k}\sum_{i=1}^k h^{tr_i}_j $ and $\sigma^{TR1}_j = \sqrt{\frac{\sum_{i=1}^k \big(h^{tr_i}-\mu^{TR1}_j\big)^2}{k}}$},$\forall j $\\

2. Estimate Gaussian distributions $\{\mathcal{N}(\mu^{TR1}_j,\sigma^{TR1}_j)\}$, $\forall j $\\

3.\hbox{Compute $\mu^{TR2_m}_j = \frac{1}{N_{tr}-k-1}\sum_{\substack{{i=k+1}\\i \neq m}}^{N_{tr}} h^{tr_i}_j $ and} \hspace*{0.3cm}\hbox{$\sigma^{TR2_m}_j = \sqrt{\frac{ \sum_{\substack{{i=k+1}\\i \neq m}}^{N_{tr}} \big(h^{tr_i}-\mu^{TR2_m}_j\big)^2 }{N_{tr}-k-1}}$}$, \forall j $\\

4. Estimate Gaussian distributions $\{\mathcal{N}(\mu^{TR2_m}_j,\sigma^{TR2_m}_j)\}$, $\forall j $\\

5a. Considering $\mathcal{N}(\mu^{TR2_m}_{j} ,\sigma^{TR2_m}_{j}), \forall j $ and $\forall tr_i $, 
compute:\\\hspace*{0.3cm} $$\scriptsize P^{TR2_m}_{i_{j}} \doteq \mathbb{P}(x \in[h_{j}^{tr_i}-\sigma^{TR2_m}_j,h_{j}^{tr_i}+\sigma^{TR2_m}_j]|r_j),$$\\\hspace*{0.3cm}$CP^{TR2_m}_{i_{j}}\doteq 1-P^{TR2_m}_{i_{j}}$\\

5b. Compute the entropy:
\begin{equation*}
    H(tr_i)= - \sum_{j=1}^{N_r} [P^{TR2_m}_{i_{j}}\cdot log(P^{TR2_m}_{i_{j}})+CP^{TR2_m}_{i_{j}}\cdot log(CP^{TR2_m}_{i_{j}})],\; \forall tr_i
\end{equation*}

6a. Considering $\mathcal{N}(\mu^{TR1}_{j} ,\sigma^{TR1}_{j}), \forall j$ and $\forall tr_i$ compute:\\ $$P^{TR1}_{i_{j}} \doteq \mathbb{P}(x\in [h_{j}^{tr_i}-\sigma^{TR1}_j,h_{j}^{tr_i}+\sigma^{TR1}_j]|r_j),\hspace{0.1CM}CP^{TR1}_{i_{j}} \doteq 1-P^{TR1}_{i_{j}}$$\\

6b. Compute the \Saracommdue{weighted conditional} entropy, $\forall tr_i$:
\begin{equation*}
    H(tr_i|TR1)=- \sum_{j=1}^{N_r} \text{\Saracommdue{$\gamma_j^{tr_i}$}}\cdot[P^{TR1}_{i_{j}}\cdot log(P^{TR1}_{i_{j}})+CP^{TR1}_{i_{j}}\cdot log(CP^{TR1}_{i_{j}})],
\end{equation*}
\Saracommdue{where $\gamma_j^{tr_i}  \doteq \frac{P^{TR2_m}_{i_{j}}}{P^{TR1}_{i_{j}}}$}


7. Compute the average entropies:

\begin{equation*}
   H(TR2_m)= {1\over{N_{tr}-k-1}}\sum_{\substack{{i=k+1}\\i \neq m}}^{N_{tr}} H(tr_i)
\end{equation*}

\begin{equation*}
  H(TR2_m|TR1)= {1\over{N_{tr}-k-1}}\sum_{\substack{{i=k+1}\\i \neq m}}^{N_{tr}} H(tr_i|TR1)
\end{equation*}

8. Measure the \Saracommdue{similarity between $TR2_m$ and $TR1$ considering $ RBI_{TR1-TR2_m}$}:
\begin{equation*}
RBI_{TR1-TR2_m} \doteq { H(TR2_m)\over H(TR2_m|TR1)}
\label{eq:perc_info}
\end{equation*}

10. Construct the baseline range $RBI_{base}$: 

\begin{equation*}
\footnotesize
    RBI_{base} \doteq  [\min_m(RBI_{TR1-TR2_m}),\max_m(RBI_{TR1-TR2_m})]
\end{equation*}\label{eq:mibaseline}

11. Compute the norms baselines ${l_p}^{base}$ as done in Algorithm \ref{algo:weight_train}.
\end{algorithm}

\begin{algorithm}[h]
\caption{Rule-based Information at Operational Stage} \label{algo:RBI_operational}
\footnotesize
$i = 1, \ldots, N_{tr}, p=1,2, j = 1,\dots, N_r, op_i \in OP$;\\
Inputs: \hbox{ $TR1$ and $OP $ (Table \ref{tab:hits_matrix_operational}); baseline ranges $RBI_{base}$ and $l_p^{base} \;$ }\\
Output: ODD through $RBI$ and ${l_p}\;$\\
$\rule{8.5cm}{0.02cm}$\\

1. \hbox{Compute $\mu^{TR1}_j = \frac{1}{k}\sum_{i=1}^k h^{tr_i}_j $ and $\sigma^{TR1}_j = \sqrt{\frac{\sum_{i=1}^k \big(h^{tr_i}-\mu^{TR1}_j\big)^2}{k}}$, $\forall j$}\\ 

2. Estimate Gaussian distributions $\{\mathcal{N}(\mu^{TR1}_j,\sigma^{TR1}_j)\}$, $\forall j$\\

3. \hbox{Compute $\mu^{OP}_j = \frac{1}{N_{op}}\sum_{i=1}^{N_{op}} h^{op_i}_j $ and}\\\hspace*{0.3cm}\hbox{$\sigma^{OP}_j = \sqrt{\frac{ \sum_{i=1}^{N_{op}} \big(h^{op_i}-\mu^{OP}_j\big)^2 }{N_{op}}}$}, $\forall j$\\

4. Estimate Gaussian distributions $\{\mathcal{N}(\mu^{OP}_j,\sigma^{OP}_j)\}$, $\forall j$\\

5a. \hbox{Considering $\mathcal{N}(\mu^{OP}_{j} ,\sigma^{OP}_{j}),\; \forall j$ and $\forall op_i$, compute:}\\ \hspace*{0.3cm} $$P^{OP}_{i_{j}}\doteq \mathbb{P}(x\in [h_{j}^{op_i}-\sigma^{OP}_j,h_{j}^{op_i}+\sigma^{OP}_j]|r_j),\hspace{0.1cm} CP^{OP}_{i_{j}}\doteq 1- P^{OP}_{i_{j}} $$

5b. Compute the entropy:
\begin{equation*}
    H(op_i)=-\sum_{j=1}^{N_r} [P^{OP}_{i_{j}}\cdot log(P^{OP}_{i_{j}})+CP^{OP}_{i_{j}}\cdot log(CP^{OP}_{i_{j}})],\; \forall op_i
\end{equation*}

6a. Considering $\mathcal{N}(\mu^{TR1}_{j} ,\sigma^{TR1}_{j}), \forall j$ and $\forall op_i$, compute:\\\hspace*{0.3cm} $$P^{TR1}_{i_{j}}\doteq \mathbb{P}(x\in [h_{j}^{op_i}-\sigma^{TR1}_j,h_{j}^{op_i}+\sigma^{TR1}_j]|r_j),\hspace{0.1cm}CP^{TR1}_{i_{j}}\doteq1-P^{TR1}_{i_{j}}$$

6b. Compute the \Saracommdue{weighted conditional} entropy, $\forall op_i$:
\begin{equation*}
    H(op_i|TR1)=-\sum_{j=1}^{N_r} \text{\Saracommdue{$\gamma_j^{op_i}$}}\cdot[P^{TR1}_{i_{j}}\cdot log(P^{TR1}_{i_{j}})+CP^{TR1}_{i_{j}}\cdot log(CP^{TR1}_{i_{j}})],
\end{equation*}
\Saracommdue{where $\gamma_j^{op_i}  \doteq \frac{P^{OP}_{i_{j}}}{P^{TR1}_{i_{j}}}$}


7. Compute the average entropies: 
\begin{equation*}
   H(OP)= {1\over{N_{op}}}\sum_{\substack{{i=1}}}^{N_{op}} H(op_i)
\end{equation*}

\begin{equation*}
   H(OP|TR1)= {1\over{N_{op}}}\sum_{\substack{{i=1}}}^{N_{op}} H(op_i|TR1)
\end{equation*}

8. Measure the similarity between $TR1$ and $OP$ by considering $RBI_{TR1-OP}$ :
\begin{equation*}
RBI_{TR1-OP} \doteq { H(OP)\over H(OP|TR1)}
\label{eq:perc_info_op}
\end{equation*}

9. OoD detection:
\begin{center}
\textcolor{blue}{IF }  $RBI_{TR1-OP}\notin RBI_{base} $  \textcolor{blue}{THEN} flag is \textcolor{red}{$on$}\\

\textcolor{blue}{IF } $l_p(tr_i,op_j) \notin {l_p}^{base} $  \textit{ for the majority of i and j }\textcolor{blue}{THEN} flag is \textcolor{red}{$on$}\\
\textcolor{blue}{IF} \textit{\{at least one flag is \textcolor{red}{on}\}} \textcolor{blue}{THEN} $OP$ is \textcolor{red}{OoD}
\end{center}
\end{algorithm}

\section{Groupwise in operation}\label{sec:groupwise}
\subsection {Incremental technique}
The method collects a bunch of operational data before processing and classifying them as in or out of distribution. For this reason, it falls in the category of groupwise methods \cite{jiang2022revisiting,onan2022bidirectional}. Differently from pointwise, groupwise confirms a type of situation (in or out), without relying on a single point that could be a spike in a steady trend. The collection phase in operation does not imply that one would wait for new $n_s$ samples to register a new split and to provide the ODD. Splits are generated continuously, as soon as new samples are collected. Incremental techniques may be also used to accelerate the computation of statistically-based features (mean, variance, skewness and kurtosis), as in the RUL and DNS problems detailed later on \cite{ITLDNS}. Like in incremental techniques, once a new sample is available, a new (operational) bunch of $n_s$ samples is built, by adding the new sample and by disregarding the most far away point in the past (of $n_s$ positions). In turn, the bunch leads to the split collection, by computing the inherent hits on the ruleset. The process assumes a sample-by-sample incremental time window, over which the following operations are performed. A new data bunch is firstly registered, a new split is calculated and a new ODD is then derived. 
\subsection {Computational issues}
The computational speed of the bunch building process depends on how fast the data samples are collected by the system (the quantity is denoted by $\delta t_0$). The speed of the split building process depends on the time required to compute the hits of the ruleset on the bunch, namely, on the latest $n_s$ data samples ($\delta t_1$). The speed of the ODD depends on the computational time of algorithms \ref{algo:weighted_operational} and \ref{algo:RBI_operational} above ($\delta t_{a2}$ and $\delta t_{a4}$, respectively). 

The computational times of the baselines in algorithms \ref{algo:weight_train} and \ref{algo:RBI_training} are less of interest as the algorithms work at design time, in which enough computational resources are assumed to be available; they however follow similar $\mathcal{O}(\cdot)$ as their respective operational versions. On the other hand, $\delta t_{a2}$ and $\delta t_{a4}$ are of interest, as algorithms \ref{algo:weighted_operational} and \ref{algo:RBI_operational} work over the deployed ML infrastructure, for which limited computational resources may be assumed. The following considerations hold for the $\delta t$ quantities.  $\delta t_0$ is outside of the scope of the paper as it depends on the environmental conditions and on the sensing architecture of the system. $\delta t_1$ is $\mathcal{O}(n_s)$ (by assuming the time to verify a rule on a data sample a constant, independently to the complexity of the rule). By referring to the computations inherent to the metrics involved in algorithm \ref{algo:weighted_operational}, $\delta t_{a2}$ is $\mathcal{O}(N_r \cdot N_{tr})$. Analogously, $\delta t_{a4}$ is $\mathcal{O}(N_{r} \cdot N_{OP})$.

\section{Case studies}
\subsection{Datasets}
Three application scenarios are considered with the inherent datasets and relevance of the ODD problem.
\subsubsection{RUL}
The first dataset concerns damage propagation modeling for aircraft engines and is taken from the NASA repository \cite{nasa_rul}. It is an important benchmark in predictive maintenance and includes four different subsets of data ($tr_1$, $op_a$, $op_b$, $op_c$), corresponding to different machines of the same factory family. The problem is interesting in the ODD perspective because one may expect a model trained on a machine (e.g., $tr_1$) to be applicable (with limited error) to another machine (e.g., $op_a$). The features are: mean ($m$), variance ($v$), kurtosis ($k$) and skewness ($s$) of the original 23 physical quantities over time. A preliminary analysis with LLM feature ranking \cite{llmappl1} individuated the following set of 7 most important features: $s_{os2}$, $m_{Nc}$, $v_{Nc}$, $v_{phi}$, $m_{htBleed}$, $s_{htBleed}$, $m_{W31}$, whose corresponding physical quantities are outlined in table \ref{tab:easa_dataset}.

\begin{table}[h]
    \centering
    \footnotesize
    \begin{tabular}{|c|c|} \hline
       Symbol  & Description  \\ \hline
        $os2$     & Operational setting 2 \\ \hline
        $Nc$     & Physical core speed \\ \hline
        $phi$     & Ratio of fuel flow to $Ps30$ \\ \hline
        $htBleed$    & Bleed enthalpy \\ \hline
        $W31$    &  HTP coolant bleed\\ \hline
    \end{tabular}
    \caption{RUL physical quantities.}
    \label{tab:easa_dataset}
\end{table}

The target variable is the Remaining Useful Life (RUL), which represents the time before the occurrence of a fault and is binarized to assume either value `0 healthy'(RUL$>$150) or `1 fault' (RUL$\leq$150). A ML classifier through LLM predicts if the engine would come into a fault state or not; $tr_1$ constitutes the in-distribution and $\mathcal{R}_{tr_1}$ the reference ruleset.
\subsubsection{Platooning}
The second dataset (platooning \cite{platooning}) addresses collision avoidance in vehicle platooning, which is one of the most celebrated application in autonomous driving. A group of vehicles is interconnected via wireless, based on the Cooperative Adaptive Cruise Control \cite{cooperative}. The behavior of the platooning system is synthesised by the physical quantities pointed out in table \ref{tab:platooning_dataset}. The physical quantities correspond to the features of the problem, which identifies potential collision in advance after a sudden brake.

\begin{table}[h]
    \centering
    \footnotesize
    \begin{tabular}{|c|c|} \hline
       Symbol  & Description  \\ \hline
        $N$     & Number of vehicles \\ \hline
        $F0$     & Breaking force applied by the leader\\ \hline
        $PER$     & Probability of packet loss \\ \hline
        $d0$    & Initial mutual distance between vehicles \\ \hline
        $v0$    &  Initial speed\\ \hline
        $d$    &  Communication delay in the inter-connection of vehicles\\ \hline
    \end{tabular}
    \caption{Platooning features.}
    \label{tab:platooning_dataset}
\end{table}

We consider two datasets: in the first one (LOW) the communication delay $d$ parameter is bounded by $0.4$ s; in the second one (HIGH), $d$ is larger than that threshold. As in the RUL case, we set a training domain: $tr_{LOW}$ (in-distribution) as well as the reference ruleset $\mathcal{R}_{tr_{LOW}}$. A typical ODD problem is thus posed (between LOW and HIGH) as $d$ has a significant impact on performance. The ODD has here a safety preserving role as it recognizes if the delay in operation is larger than the one in training. The algorithms are however not aware that delay is the key for the datasets differentiation and understand the ODD through the operational hits on $\mathcal{R}_{tr_{LOW}}$. 
\subsubsection{DNS}
The third dataset (DNS) deals with a DNS tunneling detection problem \cite{aiello2015dns}. The aim is detecting the presence of Domain Name Server intruders by an aggregation-based traffic monitoring. Silent intruders and quick statistical fingerprints generation make the tunneling detection a hard task. Table \ref{tab:dns_dataset} shows the physical quantities of the problem.

\begin{table}[h]
    \centering
    \footnotesize
    \begin{tabular}{|c|c|} \hline
       Symbol  & Description  \\ \hline
        $q$     & Size of a query packet \\ \hline
        $a$     & Size of an answer packet\\ \hline
        $Dt$    & Time interval intercurring between query and answer \\ \hline
    \end{tabular}
    \caption{DNS tunneling physical quantities.}
    \label{tab:dns_dataset}
\end{table}

Again as in the RUL case, mean ($m$), variance ($v$), kurtosis ($k$) and skewness ($s$) are extracted over the time series of the system, thus leading to 12 features. The target variable is a binary label denoting the `presence' or `absence' of a tunneling attack. Two reference datasets are as follows: the first one considers a tunneled peer-to-peer (p2p) application, that is the training (in-distribution) domain $tr_{p2p}$ (with $\mathcal{R}_{tr_{p2p}}$ as the reference ruleset), and the second refers to the tunneled secure shell (ssh) application, which is the operational setting $(op_{ssh})$. The ODD here is of interest once ssh is used in operation under the trained p2p model. It is a quite realistic situation in cybersecurity as not all attack configurations may be anticipated at design time.

\section{Results}\label{sec:results}
The first two subsections deal with understanding the ranges of the metrics in OoD conditions. The baseline ranges are reported in the first row of all the tables and represent \Saracomm{the reference to infer possible OoD in operation.} An even partial overlap between ranges in training and operation leads to a missed detection, i.e., a false negative (FN). A false positive (FP) consists of a wrong ODD for a training (in-distribution) bunch of samples. Secondly, False Negative Rate (FNR) and False Positive Rate (FPR) are reported in subsection \ref{sec:comparison}. A\Saracomm{n operational} sample of table \ref{tab:hits_matrix_alg_1_op} or table \ref{tab:hits_matrix_operational} constitutes a FN in case no OoD is declared; a sample \Saracomm{(column)} of table \ref{tab:hits_matrix_alg_1_train} constitutes a FP in case OoD is declared. Tables are built as follows. 
\Saracomm{Each column refers to a bunch of $n_s$=5000 samples, with values reflecting the number of hits for the reference ruleset.}
We consider $N_{tr}$=50 and $N_{op}$=1 in table \ref{tab:hits_matrix_alg_1_op} while $N_{tr}$=50 with $TR1= \{tr_1, \ldots, tr_{39}\}$, $TR2 = \{tr_{40},\ldots, tr_{50}\}$ and $N_{op}$=10 in table \ref{tab:hits_matrix_operational}. 
The total repetitions of the experiments for computing FPR and FNR is 2500. The section ends with subsection \ref{sec:groupwiseresults} and outlines incremental groupwise detection in operation. 

\Saracomm{Example code and data for the experiments are available at the following link: \url{https://github.com/giacomo97cnr/Rule-based-ODD}.}

\subsection{$N_{op}=1$} \label{subsec:res_Nop_1}
Tables \ref{tab:norms_rul_1} and \ref{tab:norms_rul_2} show norms, $\mu I$ and $W\mu I$ ranges over the RUL datasets. The robustness of algorithm \ref{algo:weighted_operational} is validated by the fact that all OoD ranges are significantly far away from the training baselines. The values with the norms are closer to the respective training baselines with $op_b$ than with $op_a$ and $op_c$. This is an important indication about the similarity of in ($tr_1$) and out ($op_b$) distributions. Coming back to the EASA introductive figure, it happens that $op_b$ lies in the yellow zone. Namely, the model trained on $tr_1$ is good on $op_b$ data, \Saracomm{with FPR=18\% and FNR = 27\%, which is quite close to the in-distribution performance (training and test on separate $tr_1$ samples): FPR = 18\% and FNR = 22\%}\footnote{The mentioned FNR and FPR refer to the original RUL problem, namely, they represent the errors in fault prediction.}. On the other hand, $op_a$ and $op_c$ lie in the orange zone (i.e., the $tr_1$ model is not good anymore on $op_a$ and $op_c$, \Saracomm{being FPR = 0.03\%, FNR = 99.86\% and FPR = 0.04\%, FNR = 99.88\%, respectively})\footnote{Large FNRs here may even lead to the red zone of EASA picture (catastrophic event), depending on system resilience to wrong autonomous decisions.}. \Saracomm{When the $tr_1$ model is tested on $op_b$, a good balance between FPR and FNR is still achieved; the same does not hold for $op_a$ and $op_c$, which are far away from the $tr_1$ baseline.} 

The rationale behind the $tr_1$ and $op_b$ proximity is beyond the knowledge of the authors (one may argue about some mechanical similarity of the respective engines), but inferring such proximity through ODD is quite an important achievement. In this perspective, the method should use all the metrics jointly to provide both ODD and a measure of the distance of the in and out distributions. 

\begin{table}[h]
\footnotesize
    \centering

    \begin{tabular}{|c|c|c|c|c|} \hline
        Couples & $l_1$ & FNR ($l_1$) & $l_2$ & FNR ($l_2$)\\ \hline
        $tr_1-tr_1$     & [0.09, 0.20] &  & [0.02, 0.05] &\\ \hline
        $tr_1-op_a$     & [3.16, 3.26] & 0\% &[1.05, 1.06]& 0\% \\ \hline
        $tr_1-op_b$     & [1.16, 1.40]& 0\% & [0.30, 0.34]& 0\%  \\ \hline
        $tr_1-op_c$     & [3.16, 3.26] & 0\% &[1.05, 1.06]& 0\% \\ \hline
    \end{tabular}

    \caption{Algorithms 1 and 2: RUL. Norms.}
    \label{tab:norms_rul_1}
\end{table}

\begin{table}[H]
\footnotesize
    \centering
\begin{adjustwidth}{-0.7cm}{}
    \begin{tabular}{|c|c|c|c|c|c|} \hline
        Couples & $\mu I$ & FNR ($\mu I$) & $W\mu I$ & FNR ($W\mu I$) \\ \hline
         $tr_1-tr_1$   & [0.707,1.442] &  &[0.023, 0.045]& \\ \hline
         $tr_1-op_a$      &[0.019,0.027]  & 0\% &[0.291, 0.297]& 0\%  \\ \hline
         $tr_1-op_b$     &[0.182,0.278]  & 0\% &[0.159, 0.179]& 0\% \\\hline
          $tr_1-op_c$      &[0.019,0.027]  & 0\% &[0.290, 0.297]& 0\%  \\ \hline
    \end{tabular}
\end{adjustwidth}
    \caption{Algorithms 1 and 2: RUL. $\mu I$ and $W\mu I$. }
    \label{tab:norms_rul_2}
\end{table}

\begin{table}[H]
    \footnotesize
    \centering
    \begin{adjustwidth}{-0.7cm}{}
    \begin{tabular}{|c|c|c|c|c|} \hline
       Couples  & $l_1$& FNR ($l_1$) & $l_2$& FNR ($l_2$) \\ \hline
        $tr_{LOW}-tr_{LOW}$     & [0.02, 0.12] && [0.01, 0.04]  &\\ \hline
        $tr_{LOW}-op_{HIGH}$     & [3.80, 3.90] &0\%&[1.37, 1.39] &0\%\\ \hline
    \end{tabular}
   \end{adjustwidth}
    \caption{Algorithms 1 and 2: platooning. Norms.}
    \label{tab:norms_platooning_1}
\end{table}

\begin{table}[H]
    \footnotesize
    \centering
       \begin{adjustwidth}{-0.7cm}{}
    \begin{tabular}{|c|c|c|c|c|c|} \hline
        Couples & $\mu I$ & FNR ($\mu I$) & $W\mu I$ & FNR ($W\mu I$) \\ \hline
         $tr_{LOW}-tr_{LOW}$   & [0.87,2.73] &  &[0.02,0.06]& \\ \hline
         $tr_{LOW}-op_{HIGH}$      &[0.04,0.97]  & 8\% &[0.51,0.73]& 0\%  \\ \hline
         
    \end{tabular}
    \end{adjustwidth}
    \caption{Algorithms 1 and 2: platooning. $\mu I$ and $W\mu I$.}
    \label{tab:mutual_platooning_1}
\end{table}

\begin{table}[H]
\footnotesize
    \centering
       \begin{adjustwidth}{-0.5cm}{}
    \begin{tabular}{|c|c|c|c|c|} \hline
       Couples  & $l_1$ & FNR ($l_1$) & $l_2$ & FNR($l_2$)  \\ \hline
        $tr_{p2p}-tr_{p2p}$     & [0.002, 0.090] &  &[0.002, 0.047]&\\ \hline
        $tr_{p2p}-op_{ssh}$     & [1.630, 1.770] &0\% &[0.820, 0.890] &0\%\\ \hline
    \end{tabular}
    \end{adjustwidth}
    \caption{Algorithms 1 and 2: DNS. Norms.}
    \label{tab:norms_DNS_1}
\end{table}

\begin{table}[H]
\footnotesize
    \centering

    \begin{tabular}{|c|c|c|c|c|} \hline
       Couples  & $\mu I$ & FNR($\mu I$) & $W\mu I$ & FNR($W\mu I$) \\ \hline
        $tr_{p2p}-tr_{p2p}$     & [1.5, 2.2]&& [0.01, 0.15]&\\ \hline
        $tr_{p2p}-op_{ssh}$     &[0, 0.7]  &0\%&[0.73, 0.96] &0\%\\ \hline
    \end{tabular}

    \caption{Algorithms 1 and 2: DNS. $\mu I$ and $W\mu I$.}
    \label{tab:norms_DNS_1_1}
\end{table}

\begin{table}[H]
    \centering
    \begin{tabular}{|c|c|c|c|c|} \hline
        Couples & $l_1$ & $l_2$ & RBI & FNR\\ \hline
        $tr_1-tr_1$     & [0.12, 0.19] & [0.02, 0.03] &[0.927, 0.944]&  \\ \hline
        $tr_1-op_a$     & [3.22, 3.24] &[1.05, 1.06] & 0& 0\%\\ \hline
        $tr_1-op_b$     & [1.24, 1.33] & [0.30, 0.33]&[0.040,0.041] & 0\%\\ \hline
        $tr_1-op_c$    & [3.22, 3.24] & [1.05, 1.06]&0& 0\%\\ \hline
    \end{tabular}
    \caption{Algorithms 3 and 4: RUL.}
    \label{tab:norms_rul}
\end{table}
\begin{table}[H]
    \centering 
    \begin{adjustwidth}{-0.5cm}{}
    \begin{tabular}{|c|c|c|c|c|c|} \hline
       Couples  & $l_1$ & $l_2$&RBI& FNR \\ \hline
        $tr_{LOW}-tr_{LOW}$     & [0.03, 0.09] & [0.01, 0.03] &[0.886, 0.926]&\\ \hline
        $tr_{LOW}-op_{HIGH}$     & [3.83, 3.90] &[1.37, 1.39] &[0.024,0.025]& 0\%\\ \hline
    \end{tabular}
    \end{adjustwidth}
    \caption{Algorithms 3 and 4: platooning.}
    \label{tab:norms_platooning}
\end{table}
\begin{table}[H]
    \centering
    \begin{tabular}{|c|c|c|c|c|} \hline
       Couples  & $l_1$ & $l_2$&RBI&FNR \\ \hline
        $tr_{p2p}-tr_{p2p}$     & [0.008, 0.050] & [0.005, 0.020]&[0.821,0.972]& \\ \hline
        $tr_{p2p}-op_{ssh}$     & [1.670, 1.730] &[0.830, 0.870] & 0&0\%\\ \hline
    \end{tabular}
    \caption{Algorithms 3 and 4: DNS.}
    \label{tab:norms_DNS}
\end{table}

As far as platooning and DNS are concerned, good performance are registered, except with $\mu I$ in platooning (the topic is discussed later through groupwise analysis and in the Appendix).

\subsection{$N_{op}>1$}\label{subsec:res_Nop_over1}
This section outlines the performance of algorithms \ref{algo:RBI_training} and \ref{algo:RBI_operational}, whose results are shown in tables \ref{tab:norms_rul}, \ref{tab:norms_platooning} and \ref{tab:norms_DNS}. $N_{op} > 1$ allows to exploit more information at the operational level and thus finer separation of the OoD from the baseline. 

\subsection{Comparison with canonical methods}
\label{sec:comparison}
\begin{table*}[htbp]
  \centering
    \begin{tabular}{|c|cc|cc|cc|cc|cc|}
    \hline
          & \multicolumn{2}{|c|}{Platooning ($tr_{LOW}-op_{HIGH}$)} & \multicolumn{2}{|c|}{DNS ($tr_{p2p}-op_{ssh}$)} & \multicolumn{2}{|c|}{RUL ($tr_1-op_a$)} & \multicolumn{2}{|c|}{RUL ($tr_1-op_b$)} & \multicolumn{2}{|c|}{RUL($tr_1-op_c$)} \\
\cline{2-11}          & \multicolumn{1}{|c}{FPR} & \multicolumn{1}{c|}{FNR} & \multicolumn{1}{|c}{FPR} & \multicolumn{1}{c|}{FNR} & \multicolumn{1}{|c}{FPR} & \multicolumn{1}{c|}{FNR} & \multicolumn{1}{|c}{FPR} & \multicolumn{1}{c|}{FNR} & \multicolumn{1}{|c}{FPR} & \multicolumn{1}{c|}{FNR} \\
    \hline
    KNN$^\dagger$  & 0.6\%   & 0.7\%   & 31\%   & 30\%   & 0\%   & 0\%   & 6.5\%  & 7.5\%  & 0\%   & 0\% \\
    \hline
    u-KNN \cite{KNN}  &$\leq5$\%     & 0.16\%  & $\leq5$\%   & 94\%  &$\leq5$ \%    & 0\%   & $\leq5$\%  &  96.3\%   &$\leq5$\%   & 0\% \\
    \hline
    SVM$^\dagger$   &0.3 \%   &0.9 \%   & 49\%   & 1.2\%   & 0\%   & 0\%   & 26\%  & 31.5\%  & 0\%   & 0\% \\
    \hline
    Random Forest$^\dagger$ & 0\%   & 0\%   & 32\%   & 37\%   & 0\%   & 0\%   & 0.8\%  & 5\%  & 0\%   & 0\% \\
    \hline

    Autoencoder  & 3.6\%   & 19.9\% &14.7\%   &61.7\%  & 4.1\%    & 0\%    &12.5\%   &46.8\% & 4.1\%    & 0\% \\
    \hline

    \hline

    \end{tabular}%
    \caption{ODD performance comparison considering the point-wise structure, in terms of FNRs and FPRs. Supervised methods are marked with $^\dagger$, the other ones are unsupervised.}
  \label{tab:FNR our method point}%

\end{table*}%

\begin{table*}[htbp]
  \centering
    \begin{tabular}{|c|cc|cc|cc|cc|cc|}
    \hline
          & \multicolumn{2}{|c|}{Platooning ($tr_{LOW}-op_{HIGH}$)} & \multicolumn{2}{|c|}{DNS ($tr_{p2p}-op_{ssh}$)} & \multicolumn{2}{|c|}{RUL ($tr_1-op_a$)} & \multicolumn{2}{|c|}{RUL ($tr_1-op_b$)} & \multicolumn{2}{|c|}{RUL($tr_1-op_c$)} \\
\cline{2-11}          & \multicolumn{1}{|c}{FPR} & \multicolumn{1}{c|}{FNR} & \multicolumn{1}{|c}{FPR} & \multicolumn{1}{c|}{FNR} & \multicolumn{1}{|c}{FPR} & \multicolumn{1}{c|}{FNR} & \multicolumn{1}{|c}{FPR} & \multicolumn{1}{c|}{FNR} & \multicolumn{1}{|c}{FPR} & \multicolumn{1}{c|}{FNR} \\
    \hline
    KNN $^\dagger$& 0\%   & 0\%   & 0\%   & 0\% & 0\%   & 0\%  & 0\%   & 0\%    & 0\%   & 0\% \\
    \hline
    u-KNN \cite{KNN} & $\leq5$\%   & 0\%   & $\leq5$\%   & 0\%   & $\leq5$\%   & 0\%     & $\leq5$\%   & 0\%      & $\leq5$\%   & 0\%   \\
    \hline
    SVM$^\dagger$   & 0\%   & 0\%    & 0\%   & 0\%  & 0\%   & 0\%  & 0\%   & 0\%   & 0\%   & 0\% \\
\hline
    Random Forest$^\dagger$  & 0\%   & 0\%  & 0\%   & 0\% & 0\%   & 0\%  & 0\%   & 0\%    & 0\%   & 0\% \\
    \hline

    Autoencoder &3.7\%   &0\%   &4\%   &0\% &3.3\%   &0\% &3.3\%   &0\% &3.3\%   &0\%   \\
    \hline

    Algorithm \ref{algo:weighted_operational} & 0\%   & 0\%    & 0\%   & 0\%  & 0\%   & 0\%    & 0\%   & 0\%  & 0\%   & 0\%  \\
    \hline
   
    Algorithm \ref{algo:RBI_operational}& 0\%   & 0\%    & 0\%   & 0\%  & 0\%   & 0\%    & 0\%   & 0\% & 0\%   & 0\%  \\
    \hline
    \end{tabular}%
    \caption{ODD performance comparison considering the group-wise structure, in terms of FNRs and FPRs.  Supervised methods are marked with $^\dagger$, the other ones are unsupervised.}
  \label{tab:FNR our method group}%

\end{table*}%
This section outlines a comparison with canonical supervised algorithms, such as K-Nearest Neighbours (KNN), Support Vector Machine with a RBF kernel (SVM) and Random Forest as well as \giac{unsupervised ones like the unsupervised KNN (u-KNN) and the Autoencoder}. \giac{ In particular, we first present the results considering the pointwise structure and then the groupwise counterpart.} Supervised algorithms exploit information about OoD data. A mix of the in and out data are considered for training supervised approaches and then a testing phase got the FPR and FNR values presented in table \ref{tab:FNR our method point}. In u-KNN we followed \cite{KNN} yet revisiting it according to the fact that we were not using images; hence, we have first split up the training domain into a training set and a test one and then we have tuned two parameters: the number of neighbours (K) and a distance threshold ($\lambda$) used to determine if test data are in-distribution or not; $\lambda$ was set in order to have a true negative rate of 95\% in the training domain. Despite including information about OoD data in their training, supervised algorithms fail the ODD \giac{and unsupervised methods work even worse}. \Saracomm{This may denote that} training and operational data are confused in the original feature space. \giac{The proposed algorithms, along with the above mentioned supervised and unsupervised techniques,} achieve better performance in virtue of looking at in and out separation in a different space, namely, through the ruleset hits in training. \giac{Thus, we repeated the same experiments considering the groupwise structure (Table \ref{tab:FNR our method group}) inducted by the usage of the rule hits; specifically, we used the rule hits as the input features (in place of the original features) and verified a perfect separation between in and out distributions with all the considered methods. Algorithms \ref{algo:weighted_operational} and \ref{algo:RBI_operational} are however still preferable for different reasons: first they are not black box methods, secondly they do not need significant tuning of critical parameters and finally for its robustness due to the usage of multiple metrics \cite{onan2018biomedical}.} In table \ref{tab:FNR our method group}, Algorithm \ref{algo:weighted_operational} still experiences some FPR as the weighted version of the mutual information is applied to a smaller portion of operational data than with Algorithm \ref{algo:RBI_operational}. Another rationale behind the sensible level of FPR comes from the declaration of OoD if at least one of the metrics registers an OoD. This minimizes FNR, but may increase FPR. Additional results (not reported here for the sake of synthesis) confirm that the algorithm is even more sensitive to FPR with values of $N_{tr}<50$. On the other hand, algorithm \ref{algo:RBI_operational} decreases also FPR (with respect to algorithm \ref{algo:weighted_operational}), in virtue of the (Gaussian) statistical filter applied to several splits of operational data.

\subsection{Incremental groupwise in operation} \label{sec:groupwiseresults}
\begin{figure*}[!h]
\centering
\subfloat[RUL $tr_1$-$op_a$ and $tr_1$-$op_b$ with $W\mu I$.]{\includegraphics[width=2.3in]{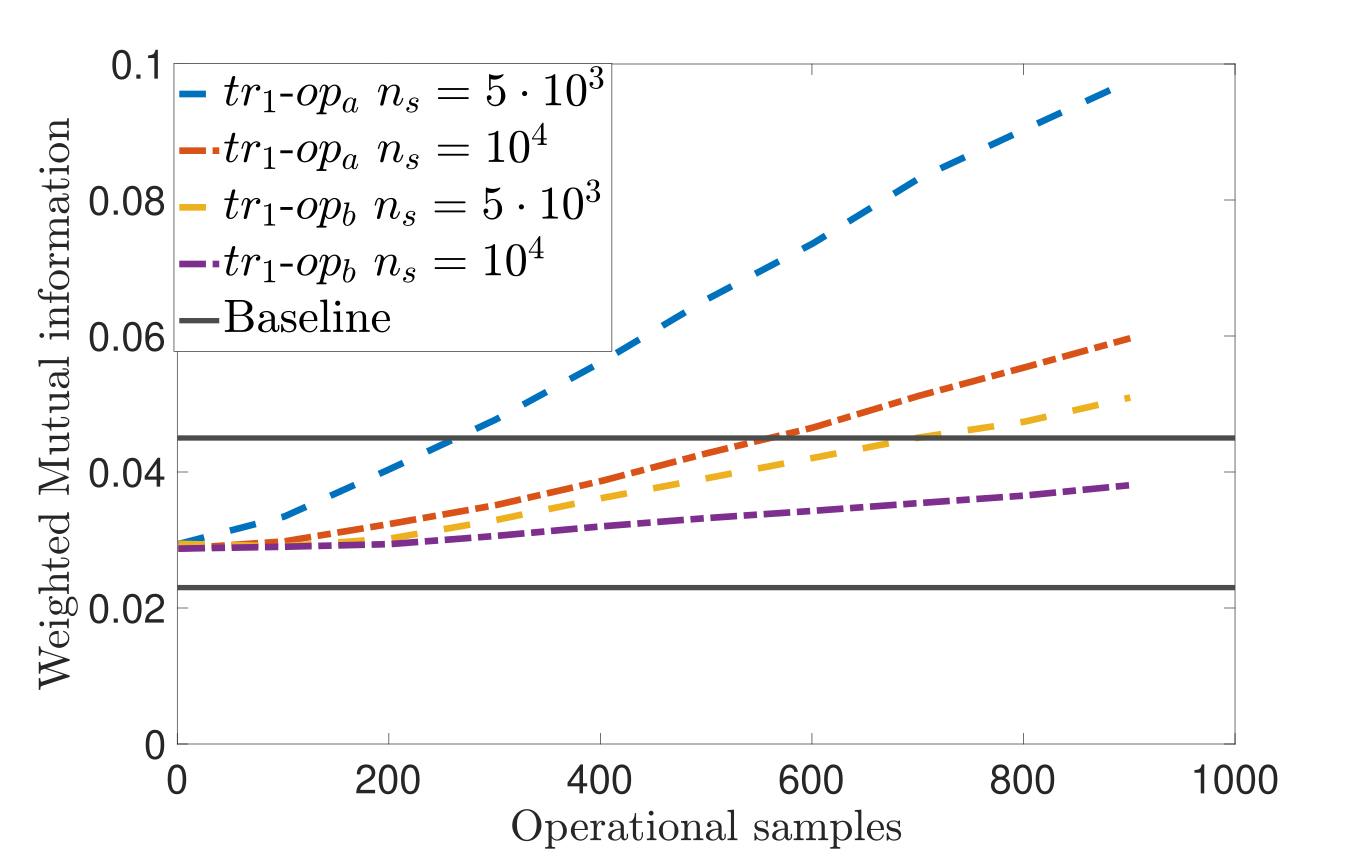}
\label{fig:rulweighted_opa}
}
\hfil
\subfloat[RUL $tr_1$-$op_a$ and $tr_1$-$op_b$ with $\mu I$.]{\includegraphics[width=2.2in]{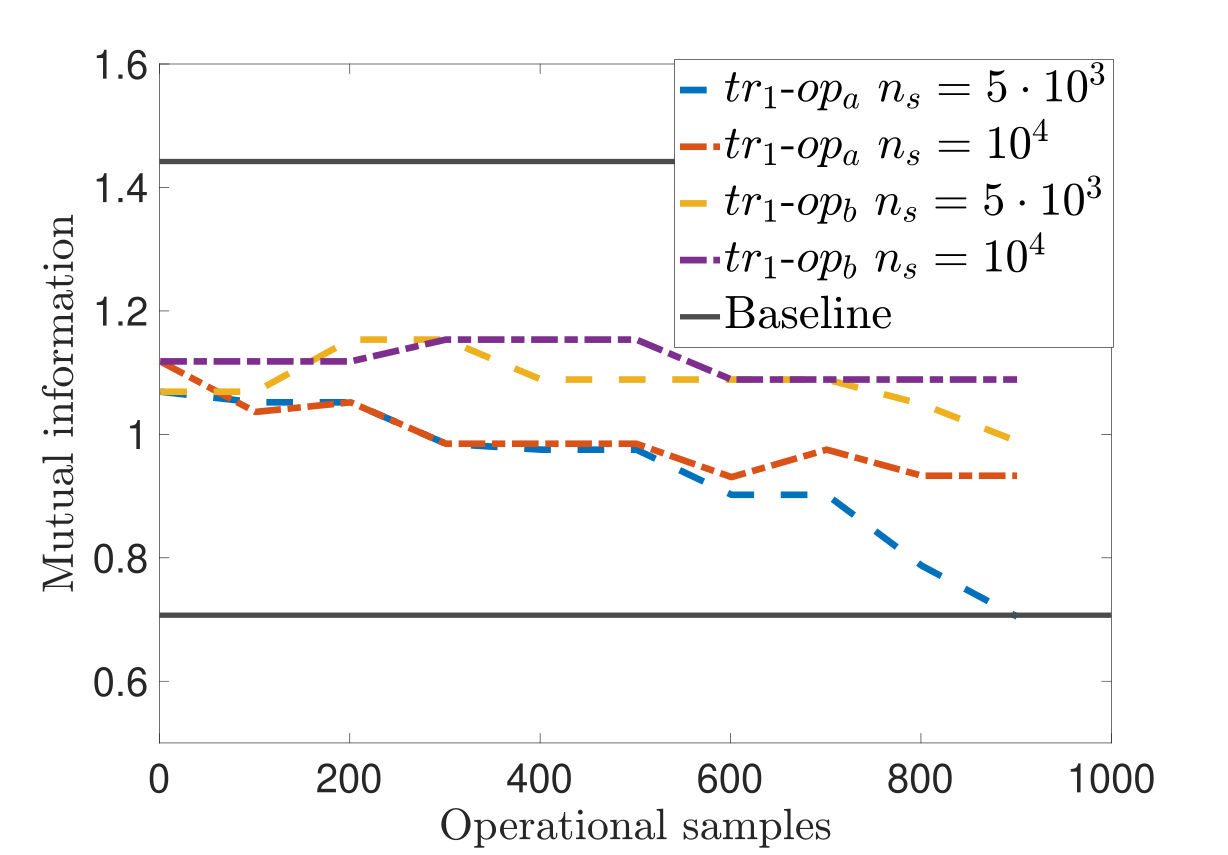}
\label{fig:rulmutual_opa}
}
\hfil
\subfloat[RUL $tr_1$-$op_a$ and $tr_1$-$op_b$ with $l_1$ norm.]{\includegraphics[width=2.2in]{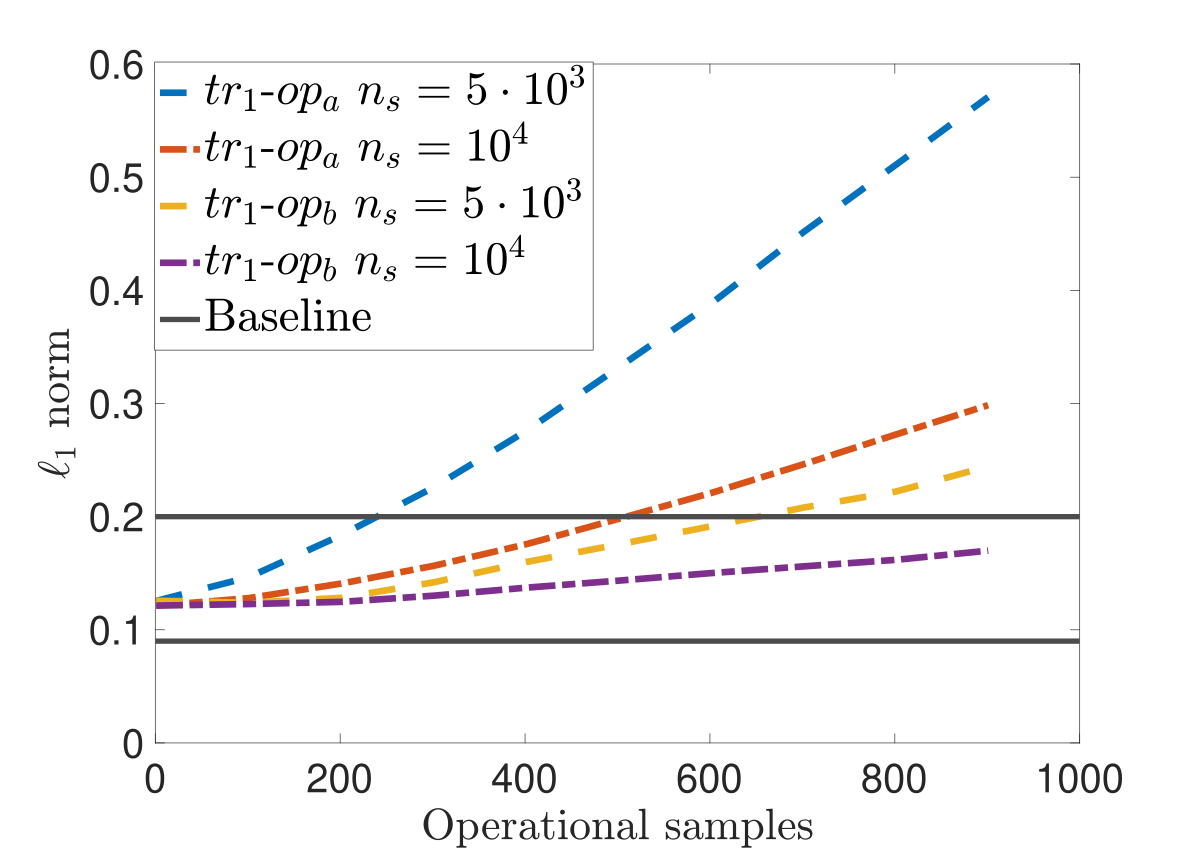}
\label{fig:rulnorm}
}
\vfill
\subfloat[Platooning with $W\mu I$.]{\includegraphics[width=2.2in]{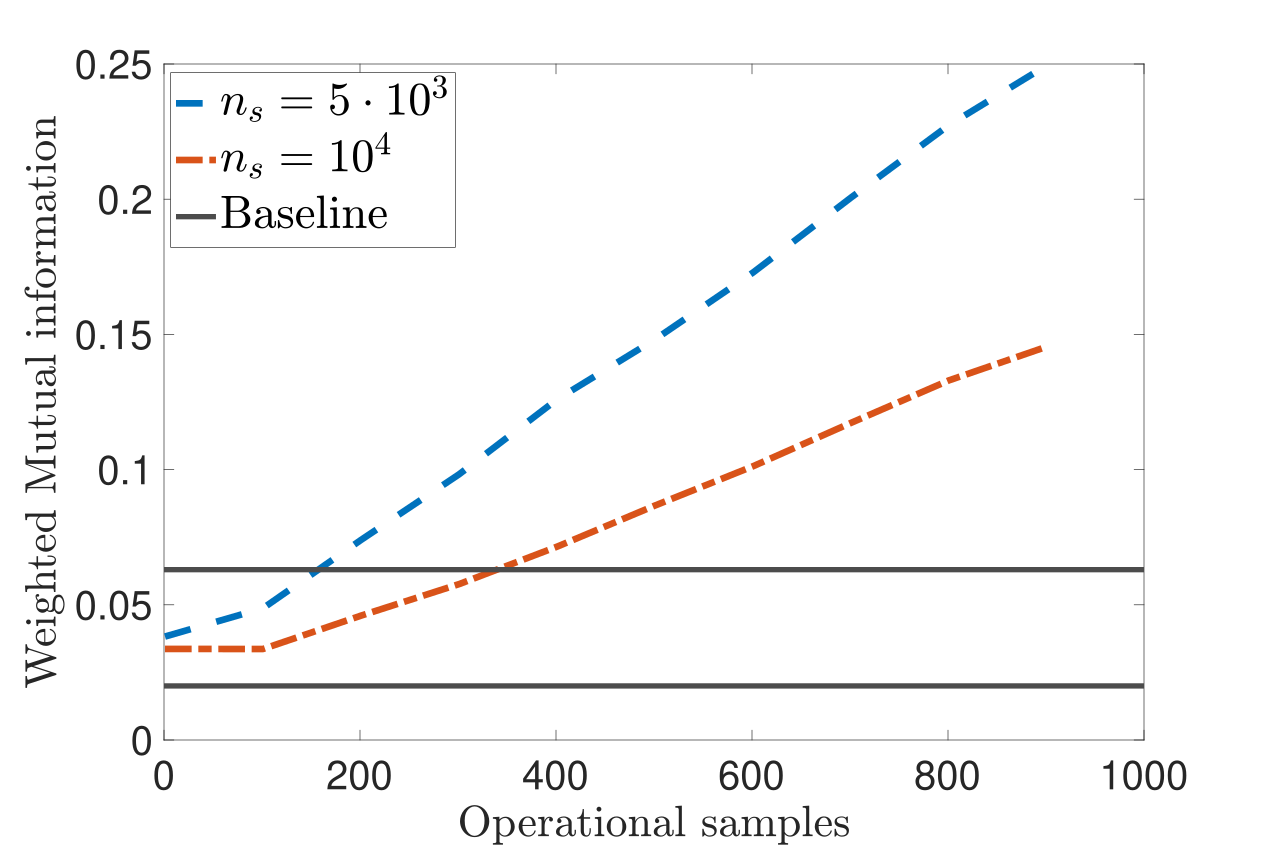}
\label{fig:platoweighted}
}
\hfil
\subfloat[Platooning with $\mu I$.]{\includegraphics[width=2.2in]{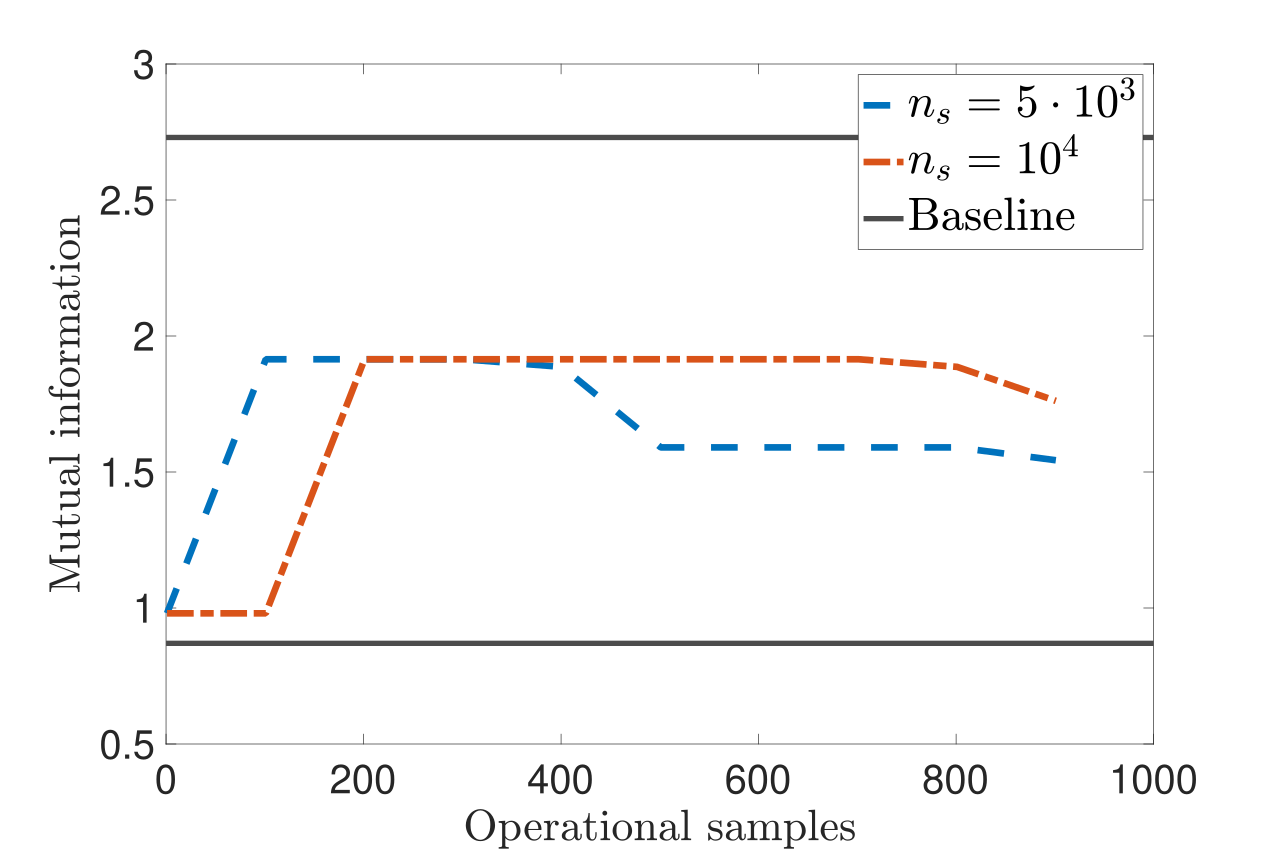}
\label{fig:platomutual}
}
\vfill
\subfloat[DNS with $W\mu I$.]{\includegraphics[width=2.2in]{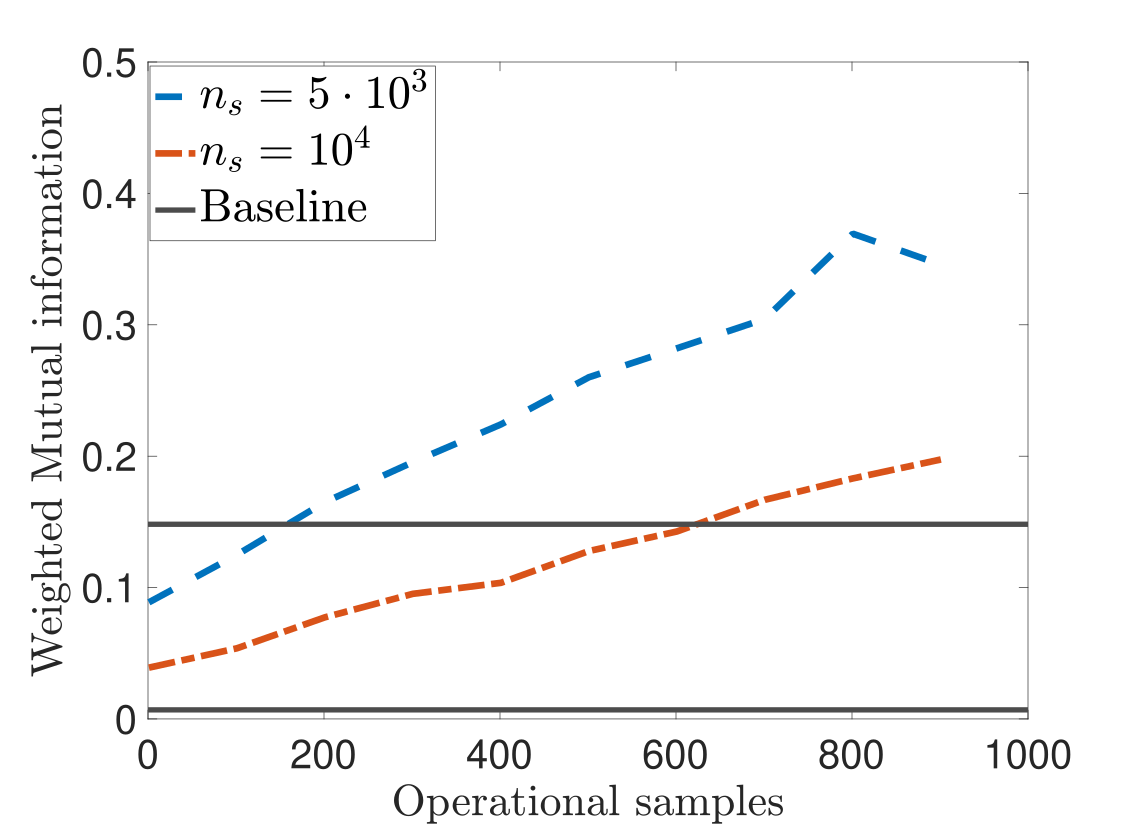}
\label{fig:dnsweighted}
}
\hfil
\subfloat[DNS $\mu I$.]{\includegraphics[width=2.2in]{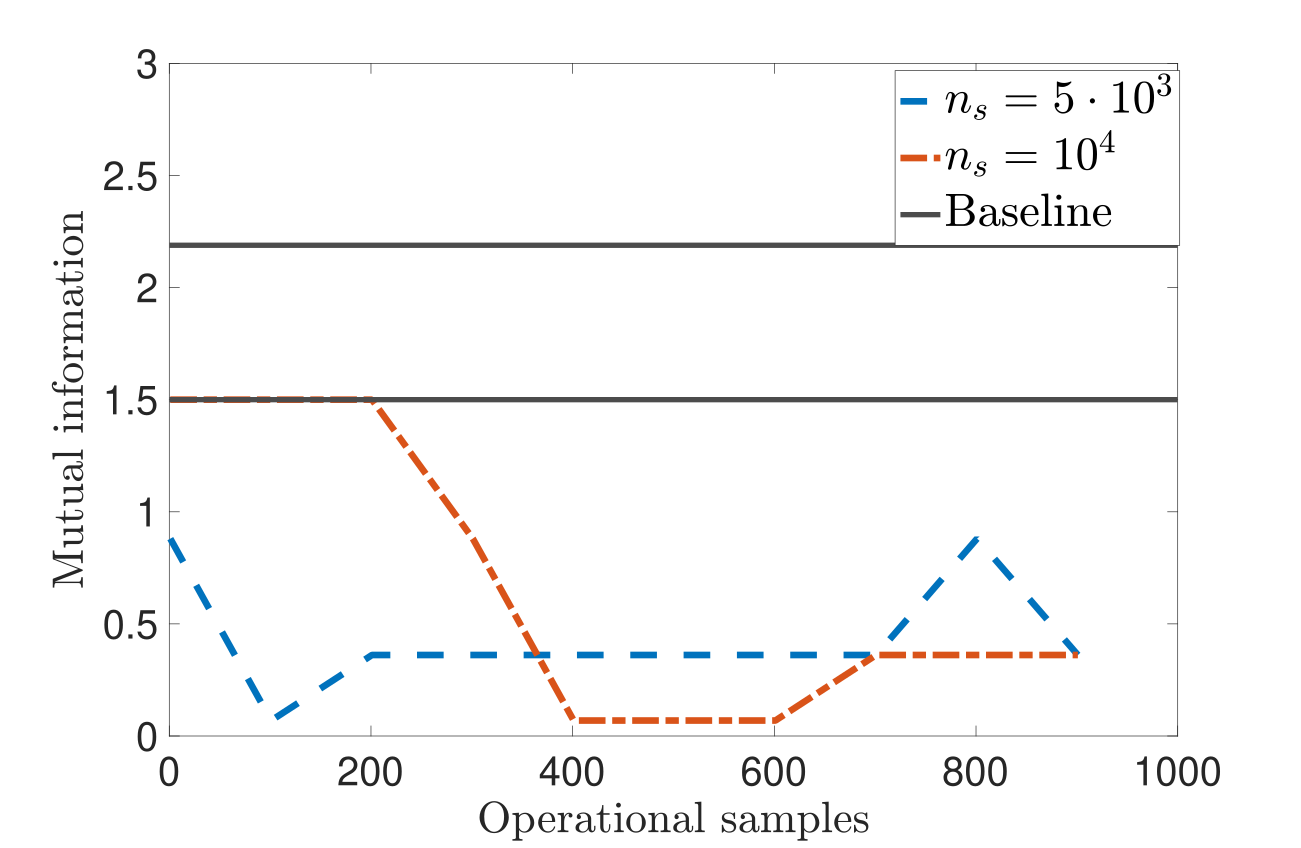}
\label{fig:dnsmutual}
}
\hfil
\caption{Incremental group-wise results}
\label{fig:groupwisePics}
\end{figure*}

By referring to section \ref{sec:groupwise}, the following experiments highlight the ODD when replacing in-distribution data with out-of, in a sample-by-sample, incremental, way. The analysis is relevant to the tracking of the OoD drift with both precision and measurement of distributions proximity. Every figure in the following contains the baseline derived at design time; the curves represent the behaviours of the metrics in operation. Increasing time windows with $n_s=5 \cdot 10^3$ and $10^4$ samples are used to emphasize the speed of the drift inference over time. The time size of the windows depends on the time granularity of the arrival of the points in operation; for this reason, the $x$-axis is not time, but it refers to the progressive identifier of the operational samples. The drift starts at time zero, that means the first operational sample derives from the OoD and previous points (of the window) are compliant with training conditions. As soon as the window collects more data (over the last $n_s$ points), it senses more information about the OoD. As to the $W\mu I$ metric, the results confirm that the shorter the window, the faster the detection. On the other hand, the $\mu I$ metric experiences a noise that can have different meanings as detailed later on. 

The following evidence arises for the case studies. In RUL, $W\mu I$ (Fig. \ref{fig:rulweighted_opa}) needs at least 200 samples to exit the baseline; this happens with the shortest window ($n_s=5000$) and with the most divergent OoD ($op_a$ with respect to $op_b$). The $l_1$ norm (Fig. \ref{fig:rulnorm}) outlines a similar behaviour. $\mu I$ (Fig. \ref{fig:rulmutual_opa}) does not trigger the expected ODD; this seems in contrast with previous results in table \ref{tab:norms_rul_2}, where ODD was successful. This is however due to the limited horizon of the figure; the curves under $n_s=5\cdot10^3$ are actually approaching the baseline and, as expected, $op_a$ reveals to be faster than $op_b$, being more divergent from $tr_1$ than $op_b$. The groupwise progression thus suggests the joint adoption of the metrics to achieve both precision ($W\mu I$) and measure of the distributions similarity ($\mu I$). 

In platooning, $W\mu I$ matches the ODD and, coherently with previous results (table \ref{tab:norms_platooning_1}), $\mu I$ is stuck in the baseline. Finally, DNS has good performance with the two metrics as well.

The difference between RUL and platooning in $\mu I$ is remarkable as it is very subtle. In the former case, $\mu I$ is sensitive to distributions similarity, still being able to slowly proceed in the ODD direction. In the latter, it experiences imprecise calculations (as shown in the appendix), thus complicating the ODD task. 


It is finally worth noting that the window of the incremental groupwise should be coherent with the design setting with $n_s=5000$. Other results may show several counterexamples in RUL with $n_s=1000$ and $tr_1-tr_1$, in which, though only points in the baseline would have been expected, many false positives take place.


\section{Conclusion and future work}
The paper deals with the identification of out-of-distribution through a distributional assumption free rule-based model. The approach also measures the proximity of in and out of distributions and is validated in challenging case studies. Future extensions comprise further testing on additional longitudinal datasets, as well as on image data. Alternative ways to the hits of the ruleset to infer in-distribution behaviour are of interest, as well as additional metrics to measure in and out of distribution divergence. 

\section*{Appendix}
\subsection*{Rationale of mutual information modification}\label{subsec:RULE}
When comparing couples of histograms, ($\mu I$) is useful to identify the dependence but it does not capture the differences among their values. Suppose we get these three  histograms $A$, $B$  and $C$ (Table \ref{tab:controex:numerico}).
\begin{table}[!h]
\caption{Example of $\mu I$}
    \centering
    \begin{tabular}{|c|c|c|c|} \hline
         &$A$&$B$&$C$\\ \hline
      $r_1$ &0.166	&0.211	&0.399 \\ \hline
      $r_2$ &0.182	&0.214	&0.387 \\ \hline
      $r_3$ &0.438	&0.387	&0.214 \\ \hline
      $r_4$ &0.424	&0.399	&0.211\\ \hline
     
    \end{tabular}
    \label{tab:controex:numerico}
\end{table}
Considering the simple $\mu I$, histograms $A$ and $B$ are dependent and so $A$ and $C$ are; but $B$ and $C$ are different (they have same values but in different positions). Since each row of the $tr$ and $op$ histograms corresponds to a rule, $\mu I$ may have a detrimental effect as the rule hits contain the information to the OoD. The correction to overcome this issue consists of weighting the probabilities (used in entropy calculations) through the average of hits differences in each rule/row; this leads to $\alpha_{i,j}$ quantities in Algorithm \ref{algo:weight_train}. The more the histograms are dependent and similar, the more $W\mu I$ goes towards zero.
Similar considerations hold for $RBI$, with $N_{op}>1$, the weights ($\gamma_j^{(\cdot)}$ quantities) are the fractions of the compared probabilities. 

\section*{Acknowledgements}
This work was supported in part by REXASI-PRO H-EU project, call HORIZON-CL4-2021-HUMAN-01-01, Grant agreement ID: 101070028. G. De Bernardi PhD is partially funded by Collins Aerospace.

\bibliographystyle{ieeetr} 
\bibliography{biblio.bib}

\begin{IEEEbiography}[{\includegraphics[width=1in,height=1.25in,clip,keepaspectratio]{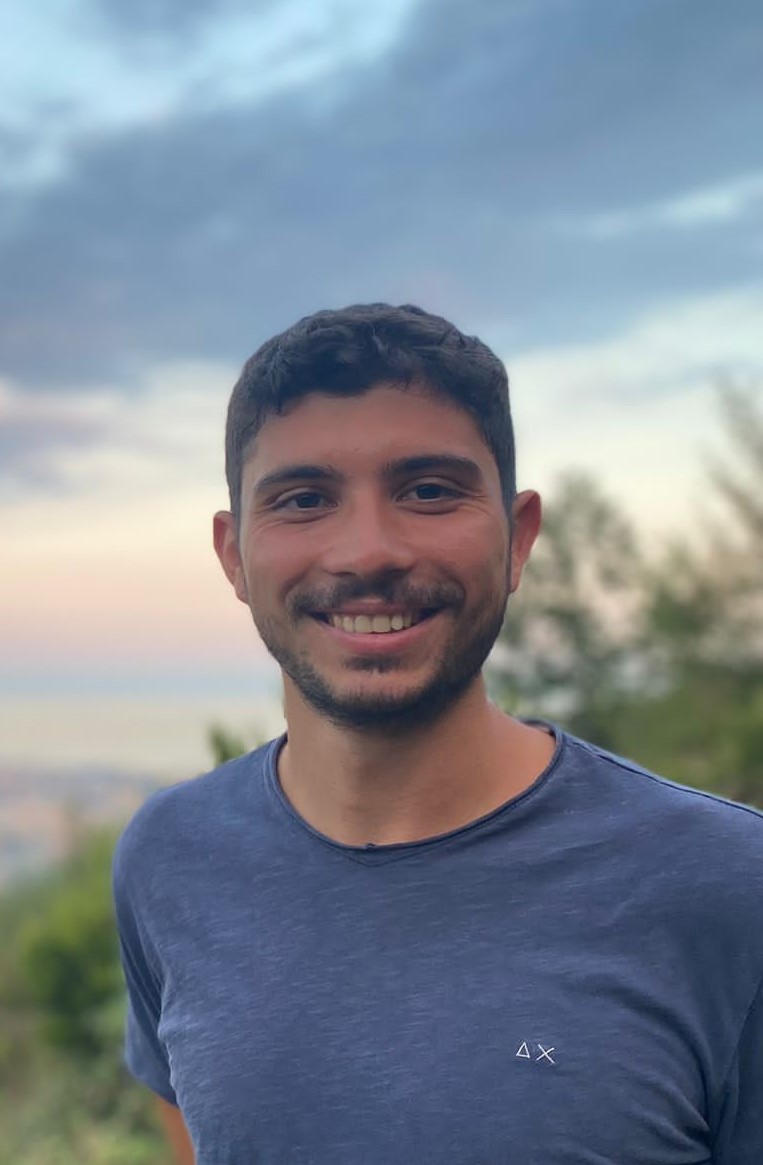}}]{Giacomo De Bernardi} 
Got his Master degree  in Mathematics at the University of Milano Bicocca on November 2021. He is currently a PhD student in the PhD programme on Trustworthy AI at university of Genoa, working at CNR-IEIIT Institute and at Collins Avionics. He works on data analytics topics from different fields, such as industry, aerospace and automotive, with specific focus on Explainable Artificial Intelligence, deep learning and machine learning methods and applications.
\end{IEEEbiography}

\begin{IEEEbiography}[{\includegraphics[width=1in,height=1.25in,clip,keepaspectratio]{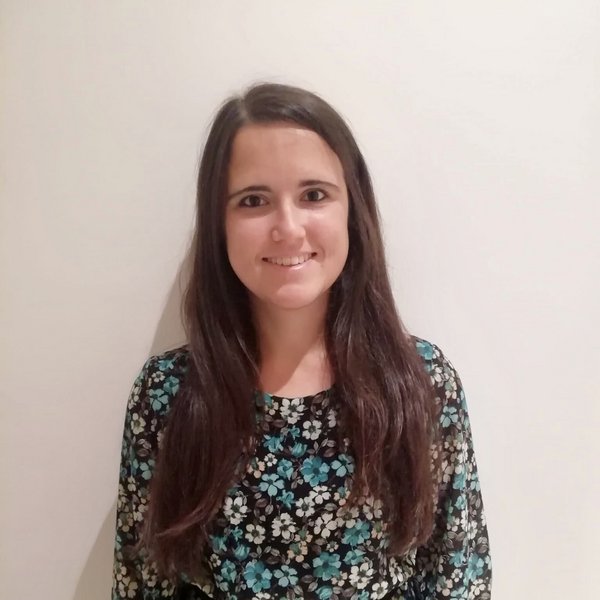}}]{Sara Narteni}
Got her M.Sc. in Bioengineering at the University of Genoa on March 2020. 
She is currently a PhD student in the italian National PhD programme on Artificial Intelligence at Politecnico di Torino, working at CNR-IEIIT Institute. She works on data analytics topics from different fields, such as industry, healthcare and automotive, with specific focus on Explainable Artificial Intelligence methods and applications.
\end{IEEEbiography}

\begin{IEEEbiography}[{\includegraphics[width=1in,height=1.25in,clip,keepaspectratio]{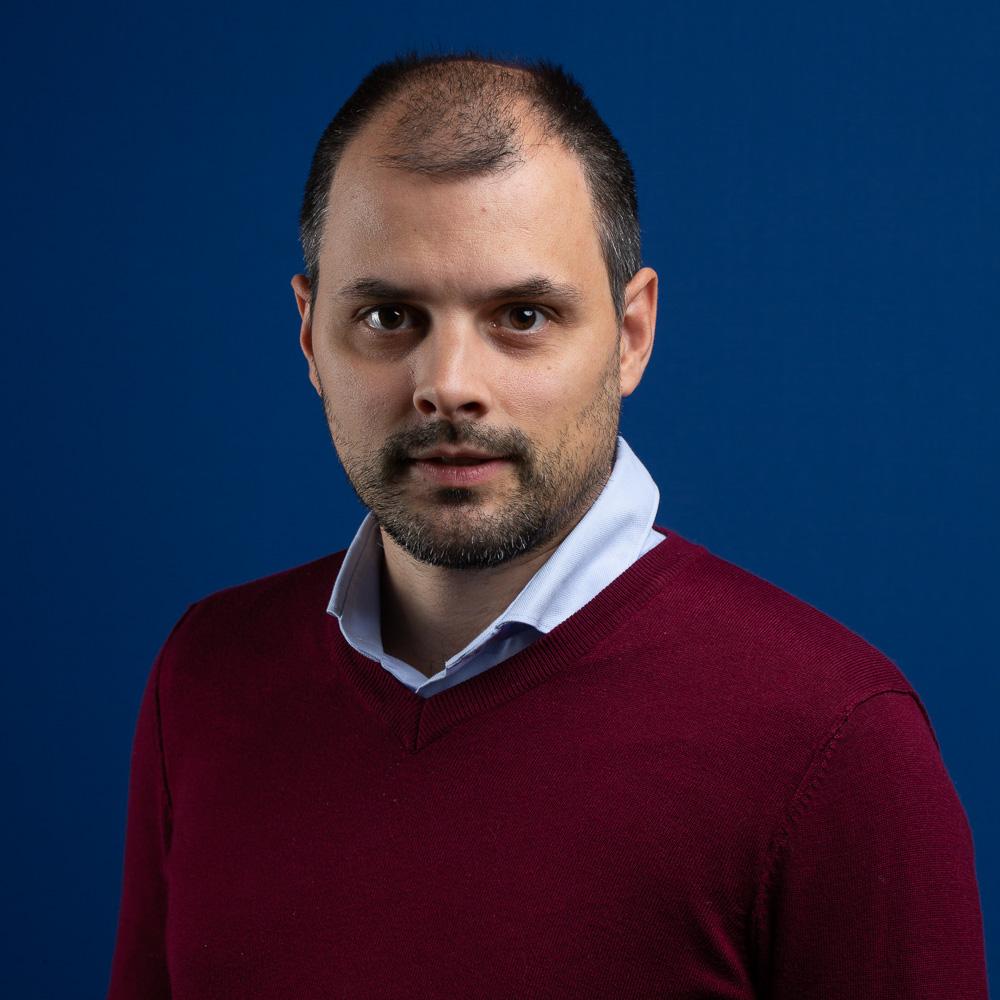}}]{Enrico Cambiaso}  Enrico Cambiaso, Ph.D in Computer Science, has a background working experience as a computer scientist, for both small and big enterprises. He is currently employed at the IEIIT institute of Consiglio Nazionale delle Ricerche (CNR), as a technologist working on cyber-security topics and focusing on the design of last generation threats. He is author of more than 50 scientific papers on cyber-security and he's been involved in several financed research projects, at national and European level.
\end{IEEEbiography}

\begin{IEEEbiography}[{\includegraphics[width=1in,height=1.25in,clip,keepaspectratio]{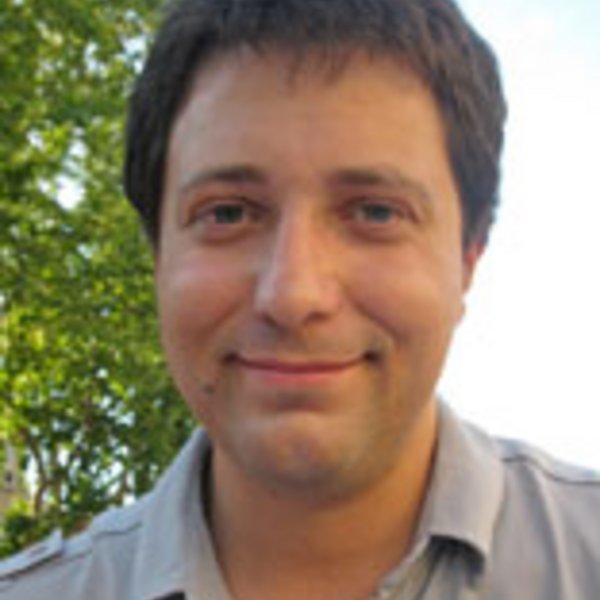}}]{Maurizio Mongelli}
obtained his PhD. Degree in Electronics and Computer Engineering from the University of Genoa in 2004. 
He worked for Selex and the Italian Telecommunications Consortium (CNIT) from 2001 until 2010. 
He is now a researcher at CNR-IEIIT, where he deals with machine learning applied to health and cyber-physical systems. He is co-author of over 100 international scientific papers, 2 patents and is participating in the SAE G-34/EUROCAE WG-114 AI in Aviation Committee.
\end{IEEEbiography}

\end{document}